\documentclass[acmsmall,screen]{acmart}
\usepackage{placeins}
\usepackage{algorithm}
\usepackage{algpseudocode}
\usepackage{amsmath}
\usepackage{listings}
\usepackage[normalem]{ulem}

\usepackage{tabularx,longtable}
\usepackage{color,colortbl}
\usepackage{multicol}
\usepackage{multirow}
\usepackage{siunitx}
\usepackage{paralist}
\usepackage{subcaption}

\usepackage{multirow}
\usepackage{tabularx}
\usepackage{adjustbox}

\newcommand{\ie}{i.e.,}
\newcommand{\eg}{e.g.,}

\newcommand{\proofparagraph}[1]{\medskip\noindent\textbf{#1.}}

\newcolumntype{C}{>{\centering\arraybackslash}X}

\usepackage{hyperref}

\usepackage[capitalize,noabbrev]{cleveref}

\graphicspath{
    {figures/}
    {figures/AddAn/}
    {figures/SupAn/}
    {figures/SupAn/7b/}
    {figures/telonew/}
}

\newcommand{\llama}[1]{LLaMA2-#1}

\definecolor{commentcolor}{rgb}{1, 0.5, 0} 

\newcommand{\eppo}{Evolutionary Pre-Prompt Optimization}
\newcommand{\EPPO}{EPPO}

\newif\ifshowchanges
\showchangesfalse
\newcommand{\change}[1]{%
  \ifshowchanges
    {\begingroup\color{red}#1\endgroup}%
    \else
    {#1}%
  \fi
}

\newif\ifshowrevisions
\showrevisionsfalse
\newcommand{\revision}[1]{%
  \ifshowrevisions
    {\begingroup\color{magenta}#1\endgroup}%
  \else
    {#1}%
  \fi
}

\author{Mathurin Videau}
\orcid{0009-0002-3672-808X}
\affiliation{\institution{Meta AI}
  \city{Paris}
  \country{France}}
\affiliation{\institution{TAU, INRIA and LISN (CNRS \& Univ. Paris-Saclay)}
  \city{Orsay}
  \country{France}}
\author{Alessandro Leite}
\orcid{0000-0002-3071-8019}
\affiliation{\institution{INSA Rouen Normandy, University of Rouen Normandy, LITIS UR 4108}
\city{Rouen}
\country{France}}
\email{aleite@insa-rouen.fr}
\author{Marc Schoenauer}
\orcid{0000-0003-1450-6830}
\affiliation{\institution{TAU, INRIA and LISN (CNRS \& Univ. Paris-Saclay)}
  \city{Orsay}
  \country{France}}
\email{marc.schoenauer@inria.fr}
\author{Olivier Teytaud} 
\orcid{0000-0001-5570-5209}
\affiliation{\institution{Thales - CortAIx}
\city{Palaiseau}
\country{France}
} 

\title{Evolutionary Pre-Prompt Optimization for Mathematical Reasoning}
\begin{document}

\frenchspacing

\begin{abstract}
Recent advancements have highlighted that large language models~(LLMs), when given a small set of task-specific examples, demonstrate remarkable proficiency, a capability that extends to complex reasoning tasks. In particular, the combination of few-shot learning with the chain-of-thought~(CoT) approach has been pivotal in steering models towards more logically consistent conclusions~\cite{wei2022chain}.
This paper explores the optimization of example selection for designing effective CoT pre-prompts and shows that the choice of the optimization algorithm, typically in favor of comparison-based methods such as evolutionary computation, significantly enhances efficacy and feasibility. Specifically,  thanks to a limited exploitative and overfitted optimization, \eppo~(\EPPO) brings an improvement over the naive few-shot approach, exceeding 10 absolute points in exact match scores on benchmark datasets such as GSM8k and MathQA. These gains are consistent across various contexts and are further amplified when integrated with self-consistency (SC)\@.
\end{abstract}


\settopmatter{printacmref=true}
\setcopyright{none}


\maketitle

\section{Introduction}
Large language models (LLMs)~\cite{brown2020language,hoffmann2022training,touvron2023llama,anil2023palm,team2023gemini} have emerged as a transformative force, demonstrating exceptional capabilities across a spectrum of tasks. However, despite their size and complexity, these models still face challenges in multi-step reasoning, particularly in tasks that require arithmetic, logic, and/or mathematical reasoning~\cite{cobbe2021training,rae2021scaling}.%

To address this limitation, recent works have focused on enhancing the reasoning abilities of LLMs. A significant advancement in this direction is the chain-of-thought~(CoT) prompting method~\cite{wei2022chain}. This approach involves guiding LLMs to articulate intermediate reasoning steps in a manner akin to human thought processes, leading to more accurate and interpretable solutions. This method has shown substantial improvements on complex tasks, including mathematics and commonsense reasoning~\cite{wei2022chain, suzgun2022challenging,lu2022learn}.%

The advancement of the CoT prompting has opened new pathways in the design of effective CoT prompts~\cite{zhou2022least,jiang2023resprompt,fu2022complexity,kojima2022large}. A crucial aspect of this research is the strategic use of complex examples in prompts for a solution. \citet{fu2022complexity} demonstrate that using such intricate examples in few-shot prompts could improve LLM performance in reasoning tasks, highlighting the importance of not only the content but also the structure of the prompts in enhancing the reasoning process. 
\change{In the context of LLMs, few-shot pre-prompting based on CoT consists of adding several examples, typically made of several steps as in standard CoT, as a pre-prompt before each prompt under consideration.}

Another effective technique in CoT prompting is the inclusion of directive phrases, such as ``let us think step by step,'' which leads to more organized reasoning generation~\cite{kojima2022large}.
Overall, these developments highlight the need to carefully select and structure prompts, as they directly affect LLM efficiency in tackling complex reasoning tasks.

In line with previous works, we investigate the effect of few-shot prompting through automated example selection. However, we use these examples as a pre-prompt for the whole downstream task at hand.
Once constructed, these few examples remain fixed for the given task (here, mathematical modeling~\cite{feigenbaum1963computers,bundy1983computer}). This differs from various in-context learning methods (see~\Cref{sec:rw}) in which prompt selection depends on each specific instance. Furthermore, since most reasoning benchmarks provide a training set that is too small for proper training~\cite{ni2022learning,liu2023tinygsm}, we focus on evolutionary algorithms, which can be comparison-based and therefore only need a few bits of feedback, paving the way for information-theoretic generalization guarantees. We analyze mathematically the risk of overfitting or exploitative behavior of such evolutionary approaches, and prove generalization bounds that do not exist for other approaches accessing fine-grained data with gradient-based methods. 

Such an \emph{\eppo} strategy (termed \EPPO\ in the following)  effectively identifies a concise set of examples (called for short {\em few-shot} in the following) that notably enhance performance when used together as a pre-prompt. Additionally, considering the current limited understanding of few-shot mechanisms~\cite{min2022rethinking}, the insights gained from the selected examples enable us to identify new ways to improve the performance of large language models, not only helping in their practical use for complex tasks, but also contributing to a better understanding of the nuances of few-shot prompting at large.

\revision{
\paragraph{\textbf{Contributions.}} The main contributions of this work are as follows:
   \begin{itemize}
      \item We introduce \eppo~(\EPPO), a comparison-based evolutionary framework for optimizing fixed few-shot chain-of-thought~(CoT) pre-prompts at the task level, framing prompt design as a principled combinatorial optimization problem.
      \item We empirically demonstrate that controlled, exploitative evolutionary search reduces overfitting in low-data regimes while outperforming brute-force example selection and classical fine-tuning.
      \item We demonstrate strong and robust empirical gains on mathematical reasoning benchmarks, achieving improvements exceeding 10 absolute exact-match points on GSM8k and MathQA, with performance that generalizes across decoding strategies and combines additively with self-consistency.
      \item We show that optimized pre-prompts induce more consistent and structured reasoning traces, revealing that prompt quality rather than prompt length is the key factor in effective and interpretable CoT reasoning.
      \item 
      We enforce these results by an information-theoretic analysis of evolutionary few-shot optimization, deriving generalization bounds that characterize the effects of data size, population parallelism, and optimization budget, and that are unavailable for gradient-based fine-tuning methods.
   \end{itemize}
}

\paragraph{\textbf{Roadmap}.} The paper is organized as follows.
\Cref{sec:methods} introduces the different components of \EPPO\ and details the overall optimization procedure. \change{It also briefly summarizes the theoretical results that bound the generalization risks of EPPO.}
\Cref{sec:xp} presents the experimental evaluation of \EPPO, with a particular emphasis on overfitting behavior and generalization.
\Cref{sec:analysis} further analyzes the internal working mechanisms of the method.
\Cref{sec:discussion} discusses the implications and limitations of the results, while \Cref{sec:rw} positions our contribution with respect to related work.
Finally, \Cref{sec:conclu} concludes the paper. %
For completeness, the theoretical analysis of the generalization risk induced by the selection process is provided in~\cref{sec:theory}, with detailed proofs of the bounds relating the optimization budget and feedback granularity to the probability of overfitting.

\section{Methods}
\label{sec:methods}
This work focuses on optimizing the performance of a given LLM on a given complex downstream task (here, mathematical reasoning) through few-shot CoT pre-prompting. 

\subsection{Datasets}
Our primary goal is to identify the most impactful small pre-prompt that can significantly boost the LLM performance on the task at hand. We will be looking for a pre-prompt composed of several prompts chosen from a much larger set, the {\em demonstration set}. We want this set of selected prompts, when used together as a pre-prompt for {\em all} further queries, to improve the efficiency and effectiveness of the LLM on the task at hand.

\change{Downstream tasks for LLMs (e.g., GSM8k, SVAMP, MathQA, among others, more in~\cref{sec:xp}) are usually defined by two sets of examples, an original {\em training set}, and a {\em test set}, both these sets being of course disjoint. The test set remains hidden throughout the EPPO process and is used only to evaluate the final solution.

To summarize, we will be dealing with three datasets made of examples of the task at hand, i.e., prompts and the corresponding expected answer from the LLM: the demonstration set is used as the source of prompts to be included in the pre-prompt; the \emph{training set} is used to compute the objective function of the optimization algorithm (see \cref{sec:optim}); and the \emph{test set}, unseen during the optimization process, is used to evaluate the best pre-prompt found by \EPPO. 
}

\begin{figure}[t]
    \centering
    \includegraphics[width=\columnwidth]{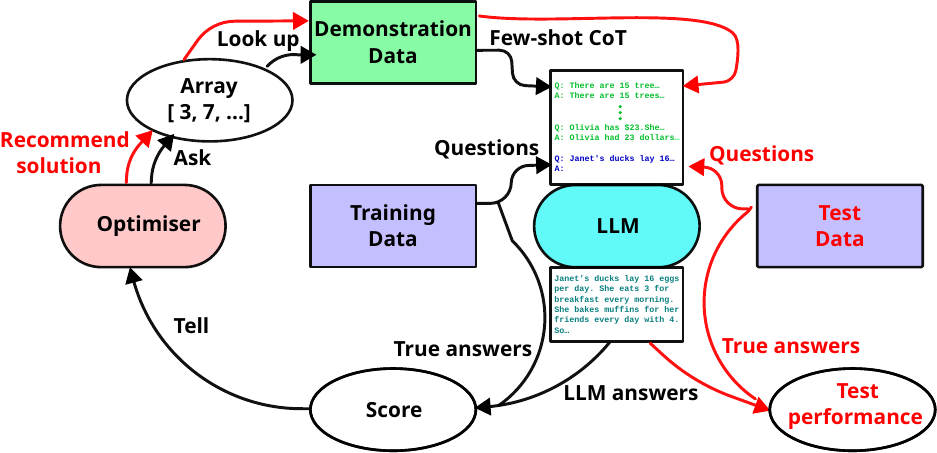}
    \caption{Overview of the proposed CoT optimization process. See \cref{sec:eppo} and \cref{algoOverview}.}
    \Description{Overview of the proposed CoT optimization process.}
    \label{fig:optim_schema}
\end{figure}

\subsection{Few-shot optimization}

For a given training set $\mathcal{T}$ and its corresponding demonstration set $\mathcal{D}$ made of CoT prompt examples, or simply {\em CoT prompts} (see \cref{sec:data_do}), we aim to find out the best performing small subset, referred to as {\em few-shot pre-prompt} in the following, or simply {\em pre-prompt}. To do so, we formulate this problem as a combinatorial optimization problem, as follows~(see \Cref{fig:optim_schema}):
%
\begin{itemize}
    \item \textbf{Representation of CoT prompts}: Each example \change{(prompt + solution)} in the demonstration set $\mathcal{D}$ is assigned a unique index in $\{1,\dots,|\mathcal{D}|\}$. This transforms the demonstration set into a lookup table. Instead of searching in $\mathcal{D}^s$, where $s$ is the size of the few-shot prompts (number of prompts) we are looking for, we search in $\{1,2,\dots,|\mathcal{D}|\}^s$. \change{Note that redundancies are possible.}
    \item \textbf{Construction of the few-shot pre-prompt}: A few-shot pre-prompt is hence represented by a (small) list of $s$ integers (typically $2\leq s \leq 16$), each representing a CoT prompt from the demonstration dataset $\mathcal{D}$. 
    \item \textbf{Optimization}: The objective is to identify the combination of integers (i.e., of CoT prompts) that, \change{when used together as pre-prompt}, maximizes the performance of the LLM for the task at hand on the training dataset. 
    This involves varying the integers of the list (and possibly the order of the prompts and the size of the list). This is amenable to black-box optimization (see~\Cref{sec:optim}). \change{The optimization loop is represented by the black arrows in \cref{fig:optim_schema}.}
    \item \textbf{Evaluation}: \change{the solution} (a few-shot pre-prompt) returned by the optimization algorithm is \change{then evaluated on the test set, not seen during the optimization. This step is represented by the \textcolor{red}{red arrows} in \cref{fig:optim_schema}.}
\end{itemize}

\subsection{Datasets construction}\label{sec:data_do}
\indent{}In the current setup, we work with \change{original} datasets that typically contain several thousand examples. However, optimizing various few-shot \change{pre-prompts} across the entire dataset is cost-prohibitive due to the high cost of running LLMs. To address this challenge, we use a subsampling strategy to create the demonstration set, from which the optimized pre-prompt will be extracted. When datasets are categorized with various difficulty levels \change{(i.e., MATH)}, we ensure a balanced approach by layered sub-sampling, i.e., uniformly sub-sampling across all categories and levels. This method ensures a diverse set of examples in the reduced dataset, which is crucial for comprehensive model evaluation.%
\\
In scenarios where the dataset lacks such detailed categorization, we pivot to a strategy based on the LLM's response uncertainty, specifically when using \llama{70B}~\cite{touvron2023llama}. We generate $n$ different answers for each example, using a predefined temperature setting $\tau$ \change{(by default $\tau=0.8$}), that influences the diversity of the model responses. To gauge confidence in an answer, we analyze the frequency of the correct answer across these multiple responses. The idea is that the more frequently a correct answer appears, the higher the confidence of the model in that answer. Based on this confidence measure, we perform uniform subsampling across all levels of uncertainty. In practice, in our experiments, we take $n = 10$ and take the same number of examples for each frequency of correct answers (in each interval between $\frac{0}{10}$, $\frac{1}{10}, \ldots \frac{9}{10}, \frac{10}{10}$). This method allows us to select a representative subset of examples that captures a wide range of the model certainty levels as a proxy for the example difficulty.\\
We apply the same strategy to construct the {\em demonstration dataset} from the remaining examples of the original training set. To further increase diversity, we additionally include correct demonstrations generated by the model using the baseline CoT. Consequently, the demonstration dataset $\mathcal{D}$ comprises both \change{the usual} hand-annotated demonstrations \change{(from GSM8k-train)} and automatically generated ones, thereby reducing the need for human intervention. \change{The proportion of generated demonstrations ranges from 0\% for Math and MathQA to 50\% for GSM8k. \Cref{tab:data_info} reports the sizes of all datasets involved.}

\subsection{Evolutionary optimization}\label{sec:optim}
The goal is to optimize an array of integers (the list of indices of \change{prompt/answer} pairs from the demonstration set). This is amenable to a classical black-box optimization setting, allowing us to use standard optimization libraries such as Nevergrad~\cite{nevergrad}. 
The simplest optimization algorithms for doing so consist of mutating one or a few variables at a time. More sophisticated methods include selecting the variables or the number of variables to modify. This last option is frequent in recent works and included in Nevergrad algorithms such as ``LogNormal''~\cite{Kruisselbrink2011lognormal} and ``Lengler''~\cite{lengler,relengler}.
In a nutshell (see \eg\ \cite{nevergrad} for details), the following optimization algorithms are used in our experiments in~\cref{sec:xp}, assuming we are optimizing an array of size $d$:

\begin{itemize}
\item The simple {\bf Discrete $(1+1)$-ES} maintains a current solution and mutates each of its variables with probability $1/d$ (repeat if no variable is mutated): the mutated point is used as a new current solution if its objective value is not worse than that of the previous one.
\item The {\bf Portfolio} method~\cite{danglehre} replaces $p=1/d$ with a uniform random choice of the mutation probability $p$ in $[0,1]$. \change{``Portfolio'' is the name used in the implementation in Nevergrad.}
\item The {\bf DoubleFastGA} method is an adaptation in Nevergrad of the Fast GA \cite{fastga} to non-binary discrete variables. 
\item In the {\bf Lengler} method, the mutation probability $p$ decreases over time, with a schedule that is mathematically derived for optimal performance on artificial test functions~\cite{relengler}. Variants include variations of the critical hyperparameters. We refer to~\cite{nevergrad} for details.
\item  The {\bf LogNormal} method~\cite{Kruisselbrink2011lognormal} uses a self-adaptation mechanism that mutates the mutation probability $p$ using a log-normal mutation before mutating the variables.
\item Some variants of these algorithms that use a crossover operator are also tried -- their name contains ``Recombining'' or ``Crossover''. The different crossover operators include \begin{inparaenum}[(a)] \item one-point crossover, \item two-point crossover, and \item uniform crossover.
We do not check for duplicates in the list after the crossover, which is unlikely to occur given the large size of the demonstration set $\mathcal{D}$ relative to the few-shot size $s$.
\end{inparaenum}

\item We also compare \EPPO\ with random search, in which all variables are repeatedly and uniformly sampled from their domains, retaining, in the end, the best-encountered solution.
\end{itemize}

The discrete algorithms discussed above are generally presented in the context of binary variables, but adapting them to categorical variables is straightforward: the mutation operator replaces the current value with a randomly drawn value from the variable's domain (excluding the current value). 

\subsection{\eppo}\label{sec:eppo}
\begin{algorithm}[htb!]
\caption{Pseudo-code for \EPPO.}\label{algoOverview}
\begin{algorithmic}[1]
\Require Random seed $\omega$, budget $b$
\Require State-transition functions $\mathrm{Init},\mathrm{Ask},\mathrm{Tell},\mathrm{Recommend}$
\Require Comparison procedure $\mathrm{Compare}$

\State $(x_0,\kappa) \leftarrow \mathrm{Init}(\omega,b,\dots)$
\Comment{$x_s$: optimizer state; $\kappa$: number of candidates per iteration}
\State $A_0 \leftarrow \emptyset$
\Comment{Archive of evaluated pre-prompts}

\For{$s = 1,\dots,b$}
    \State $(P_s, x^{\mathrm{ask}}_s) \leftarrow \mathrm{Ask}(x_{s-1})$
    \Comment{$P_s=(p_{s,1},\dots,p_{s,\kappa})$}
    \State $A_s \leftarrow A_{s-1} \cup \{p_{s,1},\dots,p_{s,\kappa}\}$\label{line:add}
    \State $i_s \leftarrow \mathrm{Compare}(P_s)$
    \Comment{$i_s \in \{1,\dots,\kappa\}$ is the index of the best candidate}
    \State $x_s \leftarrow \mathrm{Tell}(x^{\mathrm{ask}}_s, i_s, P_s)$
\EndFor

\State $r \leftarrow \mathrm{Recommend}(x_b, A_b)$
\State\Return $r$\label{line:return}
\end{algorithmic}
\end{algorithm}

A global overview of the \EPPO\ algorithm is given in \cref{fig:optim_schema}, and its pseudo-code is shown in~\cref{algoOverview}. 
The algorithm relies on a comparison-based combinatorial optimization method. Such optimization methods are often described using the so-called {\em ask-and-tell} paradigm. In this paradigm, the optimizer maintains an internal state and interacts with the evaluation process through two abstract operations. First, an {\em ask step} proposes one or more candidate solutions to be evaluated solely on the basis of the current internal state. Second, a {\em tell step} updates this internal state using feedback obtained from evaluating these candidates. Crucially, the optimizer never directly observes the evaluation process itself, but only the information returned by the {\em tell step}, which may be partial or noisy. This separation allows the same optimization logic to be reused with different evaluation mechanisms.

In \EPPO, we adopt a comparison-based variant of this paradigm: at each iteration, the {\em ask step} proposes $\kappa$ candidate pre-prompts; these candidates are evaluated by running the LLM, and the {\em tell step} receives only the index of the best-performing candidate, rather than its absolute performance value.

The internal state of the optimizer at iteration $s$ is denoted by $x_s \in \mathcal{X}$. 
The algorithm is initialized by a function
\[
(x_0,\kappa) = \mathrm{Init}(\omega,b,\dots),
\]
where $\omega$ is a random seed, $b$ is the evaluation budget, and $\kappa$ is the number of candidate pre-prompts proposed and compared at each iteration. Additional problem-specific information (\eg\ number of variables, domains, or constraints) may also be passed to $\mathrm{Init}$.

An explicit archive $A_s \subseteq \mathcal{P}$ of all visited pre-prompts is maintained. Initially, $A_0=\emptyset$. At each iteration $s=1,\dots,b$, the optimizer proposes a $\kappa$-tuple of candidate pre-prompts
\[
P_s = (p_{s,1},\dots,p_{s,\kappa})
\]
via the function $\mathrm{Ask}$, which also returns an intermediate optimizer state. All proposed candidates are added to the archive and then evaluated and compared.

The comparison procedure
\[
i_s = \mathrm{Compare}(P_s)
\]
returns the index $i_s \in \{1,\dots,\kappa\}$ of the best-performing pre-prompt among the $\kappa$ candidates. More precisely, $\mathrm{Compare}$ first evaluates each pre-prompt $p_{s,j}$ by running the LLM on the training set using $p_{s,j}$ (as illustrated in the LLM block of \cref{fig:optim_schema}), computes a scalar performance measure for each candidate, and finally returns the index of the candidate with the highest estimated performance. The optimizer does not observe the numerical scores; the comparison result $i_s$ is the only feedback signal.

The optimizer state is then updated according to
\[
x_s = \mathrm{Tell}(x^{\mathrm{ask}}_s, i_s, P_s),
\]
which emphasizes that the update depends solely on the outcome of the comparison, not on absolute performance values.

After $b$ iterations, the algorithm returns a recommended pre-prompt
\[
r = \mathrm{Recommend}(x_b, A_b),
\]
typically chosen as the best-performing element of the archive. This final pre-prompt is then evaluated on the test set (\textcolor{red}{red arrows} and \textcolor{red}{text} in \cref{fig:optim_schema}).

The explicit parameter $\kappa$ departs from the classical {\em ask-and-tell} formalism but naturally captures most evolutionary algorithms. For example, the $(1+\lambda)$-ES corresponds to $\kappa=1+\lambda$, while the $(1,\lambda)$-ES corresponds to $\kappa=\lambda$. 
Typically, \EPPO\ can be seen as a repeated A/B pre-prompt testing scenario for an online LLM, but without ensuring that the same questions are sent to both alternatives, nor that we have access to a detailed ground-truth answer for each question.  


\subsection{Generalization risk under limited feedback.}\label{sec:genrisk}

\EPPO\ selects a pre-prompt through an iterative procedure that relies on coarse feedback (\eg\ $\kappa$-ary comparisons aggregated over the full evaluation dataset), rather than on fine-grained gradient information. As a result, the returned pre-prompt is chosen from a finite set of candidates whose size depends only on the optimization budget.

Assume that for any fixed pre-prompt $p$ and any $\varepsilon>0$, the probability that the empirical loss deviates from its expected loss by more than $\varepsilon$ is bounded by $\delta_{1,\varepsilon}$. After $b$ optimization iterations with feedback arity $\kappa$, the number of pre-prompts that can possibly be returned by \EPPO\ is at most $\kappa^b$. A union-bound argument then yields
\begin{equation}
\Pr\!\left(\big| \widehat{L}(r) - L(r) \big| > \varepsilon \right)
\;\le\;
\kappa^{b}\,\delta_{1,\varepsilon},
\end{equation}
where $r$ denotes the pre-prompt returned by \EPPO.

This bound isolates the \emph{selection-induced} generalization risk of the procedure: it depends only on the feedback granularity and the optimization budget, and not on the size of the ambient pre-prompt space or on the internal complexity of the underlying language model. \revision{Detailed statements with their proofs, as well as examples of such situations, are provided in~\cref{sec:theory}.}

\section{Experimental Results}\label{sec:xp}
This section presents the experimental validation of \EPPO\ on several mathematical reasoning tasks.
\begin{table*}
    \caption{\llama{70B}: 3 baselines at the top, then 2-shots, 4-shots and 8-shots optimization and the best prompt of the 7B transferred to 70B. We observe that (i) 4-shots is best (ii) transferring the preprompt from 7b does work, though few-shots optimization is better. For MathQA, without parentheses, the exact match for the numerical answer and choice of the right option between parentheses. Across all context, the p-value for doing better than the baseline is $p < 0.02$:  the benefits of using EPPO are statistically significant.}
    \label{tab:results:70b}
    \centering\footnotesize{%
    \begin{tabularx}{\linewidth}{C|cccccccc}
    \toprule
       \multicolumn{1}{c}{\multirow{2}{*}{Prompt Type}} & \multicolumn{2}{c}{GSM8k} & \multicolumn{2}{c}{SVAMP}  & \multicolumn{2}{c}{MathQA} & \multicolumn{2}{c}{MATH} \\
      \cmidrule(lr){2-3} \cmidrule(lr){4-5} \cmidrule(lr){6-7} \cmidrule(lr){8-9}
      & \# Tokens & Acc. & \# Tokens & Acc. & \# Tokens & Acc. & \# Tokens & Acc.\\
    \midrule
      CoT & 762 & 56.8 & 762 & 73.1 & 602 & 25.4 (35.8) & 671 & 13.9\\
      Long CoT & 2838 & 66.5 & 2838 & 74.7 & 2019 & 21.9 (29.9) & 1435 & 14.2\\
      Resprompt & 2470 & 65.7 & 2470 & 71.0 & 1367 & 9.9 (13.0) & - & -\\
      \cmidrule{1-9}
      \EPPO{} 2 & 412 & 62.5 & 412 & 75.0 & 406 & 26.6 (35.6) & 843 & 14.0 \\
      \EPPO{} 4 & 1297 & \textbf{68.2} & 1297 & \textbf{78.3} & 906 & \textbf{34.1 (41.0)} & 1587 & \textbf{15.4} \\
      \EPPO{} 8 & 1651 & 67.6 & 1651 & 77.3 & 1651 & 30.5 (39.1) & 1634 & 14.7\\
      7B $\rightarrow$ 70B & 809 & 64.7 & 809 & 76.4 & 888 & 27.2 (31.4) & 641 & 14.3\\
      \bottomrule
    \end{tabularx}    
}     
\end{table*}
\subsection{Experimental settings}

\textbf{Datasets:} \EPPO\ is evaluated on four mathematical reasoning tasks. Each task comes from a different dataset, and they have heterogeneous complexity. The datasets used here are GSM8k~\cite{cobbe2021training}, SVAMP~\cite{patel-etal-2021-nlp}, MathQA~\cite{amini2019mathqa}, and MATH~\cite{hendrycks2021measuring}. GSM8k and SVAMP focus on real-world mathematical problems. While GSM8k offers a comprehensive training set, SVAMP does not have a corresponding training set. To tackle this issue, we will test the transferability of \EPPO\ by evaluating on SVAMP the same few-shot pre-prompts found for GSM8k. MathQA offers a wide range of mathematical problems of varying difficulties. In contrast, MATH poses the greatest challenge, targeting advanced mathematics typically encountered in late high school and beyond. Also, notice that MathQA has very noisy annotations and, hence, provides a test for the resilience of optimization strategies. For an in-depth understanding of each of these datasets, including specifics about their Training, Demonstration, and Test components, please refer to~\cref{tab:data_info} in~\cref{sec:datasetshyper}.

\smallskip
\noindent\textbf{Evaluation metrics:} For each dataset, our primary metric is the exact match (EM) with the ground truth. Specifically, we focus on the final output generated by the model, which is typically a numerical value, and we compare it directly with the ground truth. For the MathQA dataset, which consists of multiple-choice questions, we use two distinct EM scores. One for the numerical answer and another for the correct multiple-choice option (identified simply by its letter). This dual-scoring approach is particularly beneficial for evaluating smaller/less capable models that do not effectively correlate with the numerical answer.

\change{
\smallskip
\noindent\textbf{P-values:}
One run of \EPPO{} results in multiple evaluation loops (see \cref{algoOverview}). Thus, despite the low cost of \EPPO{} compared to a full backpropagation training procedure, such repeated experiments are still costly, and in some (though not all) of our experimental contexts, a rather small number of independent runs of \EPPO{} itself could be performed, \textbf{making comparative statistically significant tests impossible for unique contexts}. 
Nevertheless, when considering several contexts simultaneously, it becomes possible to test the statistical relevance of our findings by computing the p-values through a binomial test (Fisher's exact test), with the null hypothesis being ``the model improved by \EPPO{} does not outperform the baseline on a given context with probability $\geq \frac12$''. We refer to this measure simply by ``{\bf p-value}'' in the following, in the absence of any ambiguity.
}

\smallskip
\noindent\textbf{Language models:} We use the LLaMA family of models (i.e., LLaMA2~\cite{touvron2023llama}). A key advantage of these models is their open-source nature, which significantly supports cost-effective, reproducible research.

\smallskip
\noindent\textbf{Baselines:} %
The baselines used for comparison are:
\begin{inparaenum}[(a)]
    \item the \textbf{original CoT prompt}~\cite{wei2022chain}, as detailed in the publication and sourced from their appendix;
    \item the \textbf{Long CoT prompt}~\cite{fu2022complexity} that includes examples with more steps, targeting complex reasoning tasks. These prompts were obtained from the official GitHub repository; %
    and \item the \textbf{Resprompt approach}~\cite{jiang2023resprompt} that suggests following a reasoning graph and adding residual connections between the nodes of the graph. This technique is designed to reduce errors in multi-step reasoning tasks by LLMs. We sourced the Resprompt prompts from the paper's appendix. 
\end{inparaenum}

We do not include in-context learning such as \citep{r3p1,r3p2,r3p3,r1p1,r1p2,r1p3}, which optimize a pre-prompt on a per-example basis. Likewise, they are usually tested on heterogeneous and older models without considering mathematical reasoning tasks. Moreover, their source codes are barely available. 
 
For MATH tasks, which are not covered by any of the baselines, we took the base CoT prompt from Minerva~\cite{2022aLewkowyczMinerva}. For the Long CoT, we created prompts using the most detailed and complex answers from the training set. All these baseline prompts are 8-shot for GSM8k and SVAMP, 4-shot for MathQA and MATH. To ensure a fair comparison, we run all these baseline prompts in our own codebase. 
Our scores are consistent with the original papers, except for ResPrompt, which yields worse results on MathQA.

\smallskip
\noindent\textbf{Hyperparameters:} For each task, except for SVAMP, which lacks a training set, we create a smaller training set containing roughly 500 samples by downsampling the original one, and assemble a Demonstration dataset of about 1000 samples, as described in \cref{sec:data_do}. Precise details about each dataset can be found in~\cref{tab:data_info} in the Appendix.
In \cref{algoOverview}, the optimization algorithm $Optimiser$ is chosen among those described in \cref{sec:optim}, in their  Nevergrad implementation~\cite{nevergrad}; The budget $b$ is set to 100 for both the \llama{7B} and \llama{70B} models; The number $s$ of prompts from the demonstration set $\mathcal{D}$ is taken in $\{2,4,8, 12, 16\}$, and such pre-prompts are denoted for short ``$s$-shot pre-prompts''. 

\subsection{Results on Few-Shot Optimization}
\begin{table*}[t]
    \centering\footnotesize{
    \caption{\llama{7B}: 3 baselines at the top, then 2-shots, 4-shots, and 8-shots optimization, and the best pre-prompt of the 70B transferred to 7B. We observe that (i) 4-shots is best, (ii) no improvement when transferring the pre-prompt from 70B to 7B. Across all contexts, the p-value for doing better than the baseline is $p < 0.39$: the advantage of using EPPO is not significant here.}
    \label{tab:results:7b}
    \begin{tabularx}{\linewidth}{C|cccccccc}
    \toprule
      \multicolumn{1}{c}{\multirow{2}{*}{Pre-prompt Type}} & \multicolumn{2}{c}{GSM8k} & \multicolumn{2}{c}{SVAMP}  & \multicolumn{2}{c}{MathQA} & \multicolumn{2}{c}{MATH} \\
      \cmidrule(lr){2-3} \cmidrule(lr){4-5} \cmidrule(lr){6-7} \cmidrule(lr){8-9}
      & \# Tokens & Acc. & \# Tokens & Acc. & \# Tokens & Acc. & \# Tokens & Acc.\\
    \cmidrule{1-9}
      CoT & 762 & 14.1 & 762 & 39.6 & 602 & 10.1 (19.8) & 671 & 3.3\\
      Long CoT & 2838 & \textbf{17.5} & 2838 & 35.9 & 2019 & 9.4 (17.5) & 1435 & 3.0\\
      Resprompt & 2470 & 16.9 & 2470 & 30.4 & 1367 & 6.7 (13.8) & - & -\\
      \cmidrule{1-9}
      \EPPO\ 2-shot & 582 & 16.4 & 582 & 39.6 & 442 & 9.9 (17.6) & 460 &\textbf{4.5}\\
      \EPPO\ 4-shot & 809 & 15.5 & 809 & \textbf{41.6} & 888 & \textbf{14.7 (19.4)} & 641 & 3.9 \\
      \EPPO\ 8-shot & 2088 & 16.5 & 2088 & 39.4 & 1945 & 12.3 (16.7) & 1254 & 3.7 \\
      \EPPO\ 70B $\rightarrow$ 7B & 412 & 12.4 & 412 & 40.5 & 888 & 8.5 (14.5) & 843 & 2.6\\
      \bottomrule
    \end{tabularx}        
}    
\end{table*}

Our results, detailed in~\cref{tab:results:7b,tab:results:70b}, showcase the performance of \EPPO\ across different tasks for both the \llama{7B} and \llama{70B} models. We evaluate our approach on different $s$-shot setup with $s \in {2, 4, 8}$, to understand its effectiveness in various contexts. As shown in both tables, the results consistently indicate that \EPPO\ outperforms all baselines. This improvement is noteworthy not only for accuracy but also for efficiency, as indicated by the similar or even lower number of tokens in the few-shot pre-prompts used. This reduced number of tokens is directly reflected in lower inference costs, both in memory and in computation time. Furthermore, an important aspect of their results is the versatility of \EPPO. It is effective across models of different sizes, from the smaller \llama{7B} to the larger \llama{70B}. Interestingly, we observe more pronounced gains in performance with the larger model. This suggests that \EPPO\ scales well and can leverage the increased capacity of larger models.

\noindent\textbf{Per example complexity results:} This analysis, \change{detailed in~\cref{sec:analysis}}, examines the number of correct solutions across different levels of problem complexity for \llama{70B}: \EPPO \change{generally performs on par with or slightly better than baseline methods across a wide range of problem difficulties. The improvements are most visible on the more complex GSM8k, MathQA, and MATH problems, though these gains are modest and may partly reflect variance due to the small number of high-difficulty items. Importantly, both CoT and Long CoT also maintain strong results on these challenging instances. While EPPO was not explicitly optimized for high-complexity or multi-step problems, its discovered few-shot prompts tend to produce more uniform and often longer reasoning chains (\cref{fig:steps}), suggesting a balanced behavior between conciseness and depth.}

\noindent\textbf{Task Transfer:} As said, our approach to SVAMP, in the absence of a specific training set, involves applying few-shot pre-prompts optimized for GSM8k. This strategy leads to a notable and consistent improvement over the existing baselines across all few-shot settings. Notably, while methods like Resprompt, designed to enhance multi-step reasoning, struggle to adapt to the simpler SVAMP task, \EPPO\ demonstrates more flexibility. It successfully transfers to SVAMP and simultaneously improves performance in multi-step reasoning tasks. Further details on task transferability for \llama{70B}, using GSM8k optimized 4-shot pre-prompts, are available in~\cref{tab:task_transfert} in the appendix.

\noindent\textbf{Model transfer:} We investigate the transfer of optimized pre-prompts between two models, \llama{7B} and \llama{70B}, on the same task. We optimize pre-prompts with \EPPO\ on one of these LLMs, and test it on the other one. For \llama{70B}, we use optimized 4-shot pre-prompts, and for \llama{7B}, we choose the best from either 4-shot or 2-shot pre-prompts. The results of this experiment are presented in~\cref{tab:results:7b,tab:results:70b}. \Cref{tab:results:7b} shows the performance on the \llama{7B} model using pre-prompts optimised for \llama{70B} (indicated as 70b $\rightarrow$ 7b), and~\cref{tab:results:70b} displays the results on the \llama{70B} model of pre-prompts optimised for \llama{7B} (indicated as 7b $\rightarrow$ 70b). From these results, we observe that pre-prompts optimized for the larger \llama{70B} model do not effectively transfer to the smaller \llama{7B} model. Interestingly, the opposite scenario -- transferring pre-prompts from \llama{7B} to \llama{70B} -- led to a substantial performance increase, surpassing some baselines. Nevertheless, pre-prompts specifically optimized for one model always outperform those transferred from the other model.

\noindent\textbf{Effect of the number of shots on performance:} Each model is tested with few-shot optimized pre-prompts containing 2, 4, and 8 shots. We observe that using more than four examples does not lead to better results in the downsampled case. In fact, it appears that employing more than four examples in a pre-prompt can slightly but steadily degrade the performance of the LLM. This is a counterintuitive finding, as one might intuitively assume that providing more examples in the few-shot pre-prompt would lead to broader coverage of the task, thereby improving the LLM's ability to generate task-specific responses; hence, the investigation below in terms of overfitting.
\begin{figure}[t]
    \centering
\includegraphics[width=0.7\columnwidth]{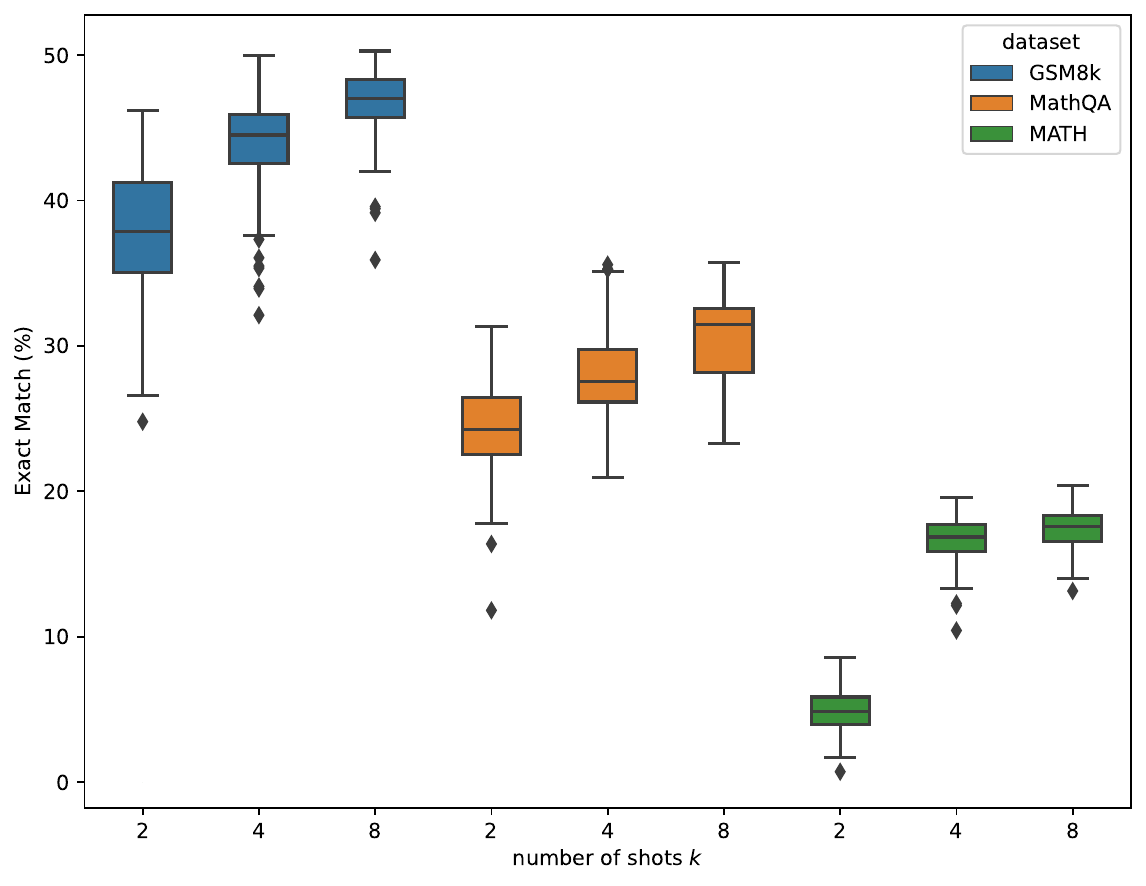}
    \caption{Comparison between 2-shot, 4-shot, and 8-shot: training score over the \EPPO\ run for \llama{70B}: Each boxplot represents the loss values observed during the whole run. The numbers represent the percentage of exact matches on the training set. Typically, for each number of shots, the bottom part (low EM) corresponds to the beginning of the run, similar to random search, and the performance of these initial few-shot increases greatly with the number of shots.
} 
    \label{fig:fewshots}%
    \Description{Comparison between 2-shot, 4-shot, and 8-shot: training score over the \EPPO\ run for \llama{70B}: each boxplot represents the loss values observed during the whole run. The numbers represent the percentage of exact matches on the training set. Typically, for each number of shots, the bottom part (low EM) corresponds to the beginning of the run, similar to random search, and the performance of these initial few-shot increases greatly with the number of shots.}
\end{figure}

\begin{table}
\small
\caption{\label{rs100} Random Search vs other optimizers, for budget $b=50$ and $100$. 
\textbf{Bold} for the best performances and \textcolor{red}{red} for the worst.
+/- denotes cases in which budget 100 outperforms (resp. underperforms) budget 50 (i.e., no overfitting). {\bf Observations}: (1) the downsampled case is more prone to overfitting (much more ``-'') (2) Random Search never overfits (consistently with predictions in \cref{sec:maths}) whereas other methods do, to the point random search is sometimes the best in the downsampled case (3) few-shot sizes 12, 16 are more prone to overfitting than 4, 8. 
}
\begin{tabular}{|c|c|c|c|c||c|c|}
\hline
 & \multicolumn{4}{|c||}{Downsampled GSM8k} & \multicolumn{2}{|c|}{Full GSM8k}\\

\hline
Few-shot size $\rightarrow$ & \multirow{2}{*}{4} & \multirow{2}{*}{8} & \multirow{2}{*}{12} & \multirow{2}{*}{16} & \multirow{2}{*}{8} & \multirow{2}{*}{12}  \\
Algorithm $\downarrow$ & &  &    &    &    &      \\
\hline
\multicolumn{7}{|c|}{Budget 50: no overfitting \& random search is weak}\\
\hline
\rowcolor{gray!25} RandomSearch     & \textcolor{red}{\bf 64.13 }  &  \textcolor{red}{\bf  64.68}  & \textcolor{red}{\bf   64.13}  & \textcolor{red}{\bf  62.16}  & 66.41  & 66.33 \\ 
Portfolio        &  64.80   & 64.80  & 65.57  & {\bf{65.54}}  & {{67.62}}  & \textcolor{red}{\bf 66.07} \\
Disc1+1        &  { 65.50} &   64.70  & 66.94  & 63.00  & {\bf{67.70}}  & {\bf{67.47}} \\
DiscLengler1+1        &  {{66.71}} &    {{66.71}}  & {\bf{67.70}}  & 65.04  & \textcolor{red}{\bf 65.73}  & 66.98 \\
RecombDiscLengler        &    {\bf{ 66.75}}  &  {\bf{66.75}}  & 65.45  &  62.62 &   65.88  & 66.48  \\

\hline
\multicolumn{7}{|c|}{Budget 100: Overfitting matters \& 
random search is competitive}\\
\multicolumn{7}{|c|}{ in the downsampled case.}\\
\hline
\rowcolor{gray!25} RandomSearch   &  66.33+   &      65.76+  & \textbf{66.33+}  & 64.44+ & \textcolor{red}{\bf 65.95} & \textcolor{red}{\bf  67.09} \\
Portfolio     & {\bf{67.46+}}   &      {\bf{67.46+}}  & \textcolor{red}{\bf 65.23-}  &  \textcolor{red}{\bf  62.62-}  & 68.00+  & 67.36+  \\
Disc1+1        & \textcolor{red}{\bf 66.18+} &    \textcolor{red}{\bf  65.30+}  & 66.07-  & {\bf{66.18+}}  & {{67.53-}} & {\bf{67.53+}}  \\
DiscLengler1+1   & 66.77+  & 66.77+  & 65.95-  & 62.01-  & {\bf{68.68+}} & 67.09+  \\
RecombDiscLengler        &     66.43-  &  66.43-  & 65.40-  &  67.55- &   67.85+  & 67.39+  \\
\hline
\end{tabular}
\end{table}

\begin{table}[t]
    \centering
    \footnotesize{%
    \caption{Self-consistency improvement over optimized prompt and CoT baseline: error in test. The model tested here is Llama-2-70B with SC over 8 generations (maj@8) and temperature $\tau = 0.6$.}\label{tab:SC}
    \begin{tabular}{l|cccc}
    \toprule
    Prompt Type & GSM8k & SVAMP & MathQA & MATH \\
    \midrule
    Cot & 56.8 & 73.1 & 25.4 (35.8) & 13.9 \\
      CoT + SC & 68.3 & 80.0 & 35.8 (44.9) & 19.1\\
      \hline
      \multicolumn{5}{|c|}{Downsampled training set}\\
      \hline
     \EPPO\ 4-shot  & 68.2 & 78.3 & 34.1(41.0) & 15.4 \\ 
      \EPPO\  4-shot + SC &        {76.9} & \textbf{85.7} & \textbf{46.7 (50.4)} & \textbf{21.3}\\
      \hline
      \multicolumn{5}{|c|}{Full training set}\\
      \hline
      \EPPO\  4-shot  + SC  & {77.1} &   82.4 & - & -\\
      \EPPO\  8-shot  + SC  & \textbf{79.1} & 85.5 & - & -\\
      \EPPO\  12-shot  + SC &       78.8       & 85.4 & - & -\\
    \bottomrule
    \end{tabular}     
}
\end{table}

\noindent\textbf{Effect of the number of shots on overfitting.} In order to understand why we get better results with a small number of shots, whereas 8 shots are usual, %
we investigate the performance of \llama{70B} after running \EPPO\ on the training set using 2, 4, and 8 shots for each task, scrutinizing the optimizer logs of the train error along optimization.
Our findings are illustrated in~\cref{fig:fewshots}, showing statistics gathered over the full \EPPO\ run,
and in~\cref{rs100} presenting test errors. 
%
    There is a noticeable and consistent performance gap between the 2-shot prompts and those with more than 4 shots in~\cref{fig:fewshots}. This indicates that increasing the number of shots from 2 to 4 yields most of the (train error) benefits.
    Also, with an increase in the number of shots, there was a corresponding increase in the median performance (\ie\ median over the \EPPO\ run) and in the performance of the worst-performing prompts (typically randomly generated at the beginning, before overfitting can take place).
    When considering the test error, \cref{rs100,fig:telonew} show that increasing the number of shots, in the downsampled case, leads to a clear overfitting. We conclude that the problem of a large few-shot is an overfitting issue, particularly visible in the downsampled case. This is confirmed by~\cref{tab:SC}, which shows good results for a larger few-shot when \EPPO{} uses the full training set (without downsampling).
%
\subsection{Experiments on the downsampled GSM8k}\label{sec:downs}
\begin{table}[t]
    \centering\footnotesize{
    %
    \caption{Comparison between exact match accuracies (without SC) obtained with different discrete optimization methods with a budget of 100 on the training set. We observe a significant performance improvement compared to random search. Red for the 2 worst performances. Additional results in \cref{rs100}.   Across all contexts, the p-value for doing better than the baseline is $p < 6.10^{-3}$, meaning that using an optimization algorithm significantly outperforms random search.}
    \label{tab:optimiser}
     \begin{tabularx}{\linewidth}{rCCC|C}
    \toprule
      & \multicolumn{3}{c}{LLaMA2-7B} & \multicolumn{1}{c}{LLaMA2-70B}\\
    \cmidrule{2-4} 
    \cmidrule{5-5}
      & \multicolumn{4}{c}{GSM8k num-shots}\\
        \cmidrule{2-5}
    \multicolumn{1}{c}{Optimizer} & 8-shot  & 4-shot & 2-shot & 8-shot \\ 
    \midrule
\rowcolor{gray!25}    RandomSearch &  \textcolor{red}{\bf 13.0} & \textcolor{red}{\bf 13.0} &  \textcolor{red}{\bf 13.3} & \textcolor{red}{\bf 46.6} \\ 
    LogNormal$(1+1)$ & \textcolor{red}{\bf 12.8} & \textcolor{red}{\bf 13.5} & \textcolor{red}{\bf 12.8} & \textcolor{red}{\bf 45.1} \\ 
    DoubleFastGA & 15.0 & 14.8 & 14.3 & 51.0 \\ 
    Portfolio$(1+1)$ & 15.5 & 14.5 & 14.8 & 49.7 \\ 
    {\textbf{Discrete$(1+1)$}} & 15.8 & \textbf{15.5} & \textbf{14.8} & \textbf{51.7} \\ 
    Lengler$(1+1)$ & 14.8 & 14.8 & 14.8 & 50.3 \\ 
    TPE & 12.5 & 13.0 & 12.0 & 45.2 \\ 
    \bottomrule
\end{tabularx}    
}    
\end{table}
    
\cref{tab:optimiser} presents a sanity-check comparison against pure Random Search of different optimization algorithms used within \EPPO, applied to the downsampled GSM8k training set (1/16 of the complete set, see \cref{sec:data_do} for details). 
We observe a gap between the train (\cref{tab:optimiser}) and test (\cref{tab:SC}, \cref{rs100}, \cref{fig:telonew}) scores of Llama-70B on GSM8k, primarily due to the downsampling procedure, which leads to the different difficulty levels being more uniformly represented in the training set than in the test set.
As said, we use the algorithms described in \cref{sec:optim} in their Nevergrad implementation~\cite{nevergrad}:  LogNormal~\cite{Kruisselbrink2011lognormal}, DoubleFastGA~\cite{fastga}, 
Portfolio~\cite{danglehre}, the basic Discrete ES, and Lengler~\cite{relengler}; We refer to ~\cite{lengler} for a theoretical analysis including the classical Discrete $(1+1)$-ES.

Most algorithms (except LogNormal) clearly outperform random search (in red), while the Discrete (1+1)-ES performs best for all small-shot sizes (in bold). Given the high computational cost of LLM inference, choosing an efficient optimizer is important for managing computational costs effectively.

\subsection{Extension to the full training set: reducing the overfitting}\label{sec:telonew}
\begin{figure}[t]
\begin{tabular}{cc}
\includegraphics[width=.40\textwidth]{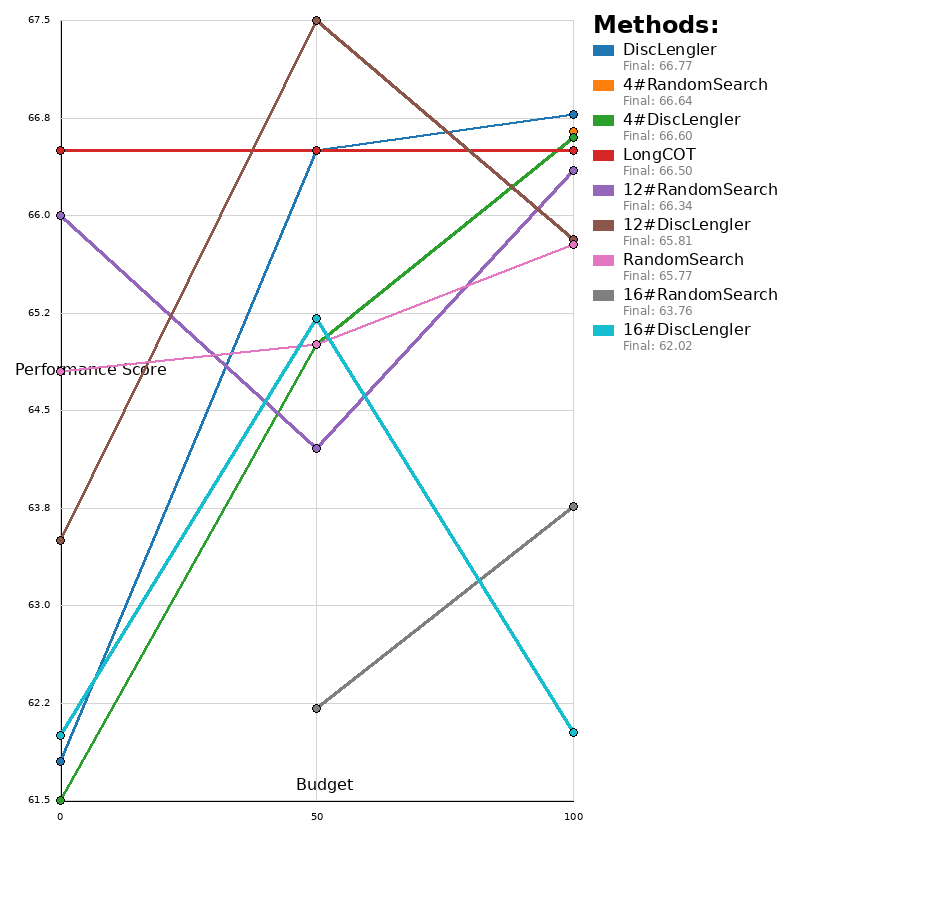} &
\includegraphics[width=.40\textwidth]{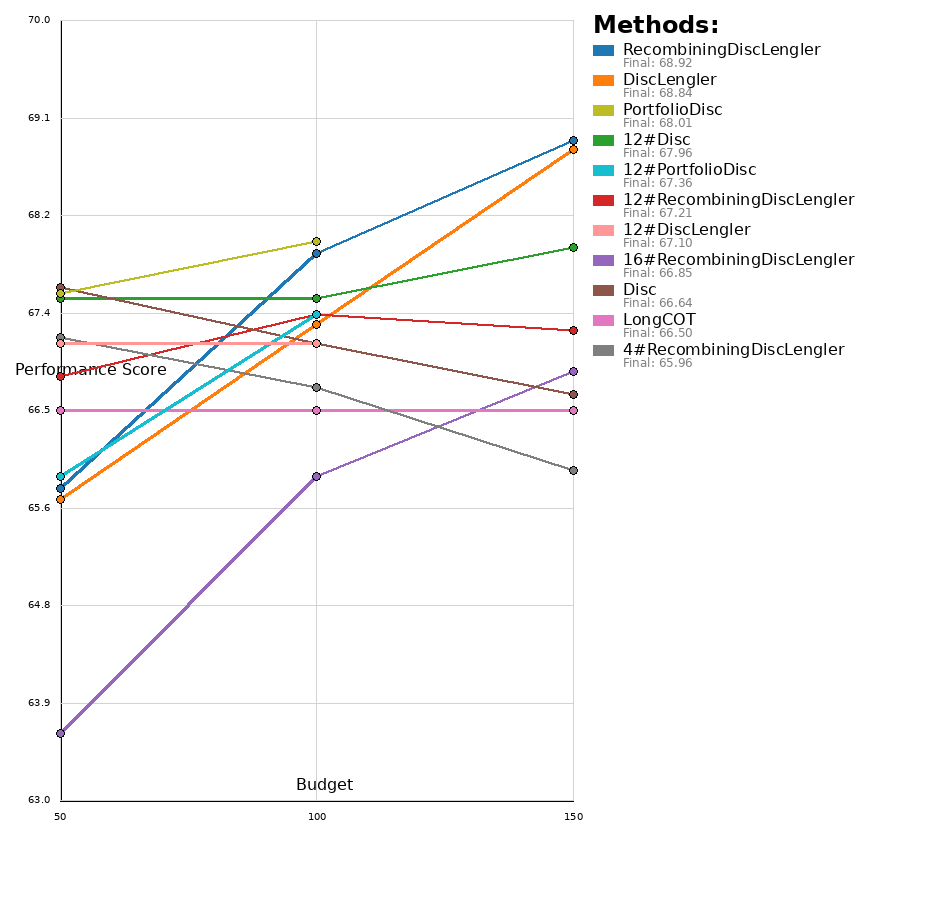}\\
4, 8*, 12 or 16 shots, & 4, 8*, 12 or 16 shots \\
Downsampled (1/16th) GSM8k  & Full GSM8k\\ 
\end{tabular}
\caption{\label{fig:telonew}Results of Llama 70B, Exact Matches on the test set without improvement by Self-Consistency. Legends are the Nevergrad algorithm names, prefixed with ``s\#'' ($^\star{}s=8$ is the default few-shot size $s$, unless mentioned). For all algorithms, $\kappa=2$, so the x-axis is exactly the number of binary comparisons between pre-prompts, i.e., budget $b$ in \cref{algoOverview}. 
\Cref{fig:svamp} presents the results of the transfer to SVAMP.\\ {\bf{Left: Downsampled GSM8k.}} For $s\in\{12,16\}$, we observe a clear overfitting:
Test performance decreases as the budget increases, consistent with the mathematical analysis. We also observe no overfitting for Random Search, which grows steadily with $s \in \{4,8,12,16\}$.  
Note that the longest run used 16 GPUs during $\simeq$ 30 hours.\\
{\bf{Right: Full GSM8k.}}  As only one bit of information is used per iteration (comparison with the best so far), we observe no overfitting until budget 150. Note that the longest run here used 160 GPUs during $\simeq$ 48 hours.}
\Description{Results of Llama 70B, Exact Matches on the test set without improvement by Self-Consistency. Legends are the Nevergrad algorithm names, prefixed with ``s\#'' ($^\star{}s=8$ is the default few-shot size $s$, unless mentioned). For all algorithms, $\kappa=2$, so the x-axis is exactly the number of binary comparisons between pre-prompts, i.e., budget $b$ in \cref{algoOverview}. 
\Cref{fig:svamp} presents the results of the transfer to SVAMP.\\ {\bf{Left: Downsampled GSM8k.}} For $s\in\{12,16\}$, we observe a clear overfitting (confirmed with more experiments in \cref{apptotal}): The test performance decreases when the budget increases, consistently with the mathematical analysis. We also observe no overfitting for Random Search, which grows steadily with $s \in \{4,8,12,16\}$.  
Note that the longest run used 16 GPUs during $\simeq$ 30 hours.\\
{\bf{Right: Full GSM8k.}}  As only one bit of information is used per iteration (comparison with the best so far), we observe no overfitting until budget 150. Note that the longest run here used 160 GPUs during $\simeq$ 48 hours.}
\end{figure}

\citet{liu2023tinygsm} noted that learning from the GSM8k training set is difficult due to its moderate size. We investigate in the present section to what extent \EPPO\ is relevant for the training set of GSM8k.
\change{The bounds on the risk of a large generalization error that we derived in \cref{sec:maths} imply that the risk of a strong overfitting is low}, thanks to the limited access of \EPPO\ to fine-grain data, but can be further limited by increasing the training set size $T$ (\Cref{eq:devh}). \cref{fig:telonew} presents experimental results validating these theoretical findings, and extending this claim with results both on one-sixteenth of the training set (downsampled GSM8k as in~\cref{sec:downs}) on the left and on the entire training set (though still using only a single scalar indicator per few-shot pre-prompt) on the right: We emphasize positive results on GSM8k with positive transfer on SVAMP with just 50 to 150 bits of information from the data.\\

\begin{figure}
\includegraphics[width=.6\textwidth,height=0.5\textwidth]
{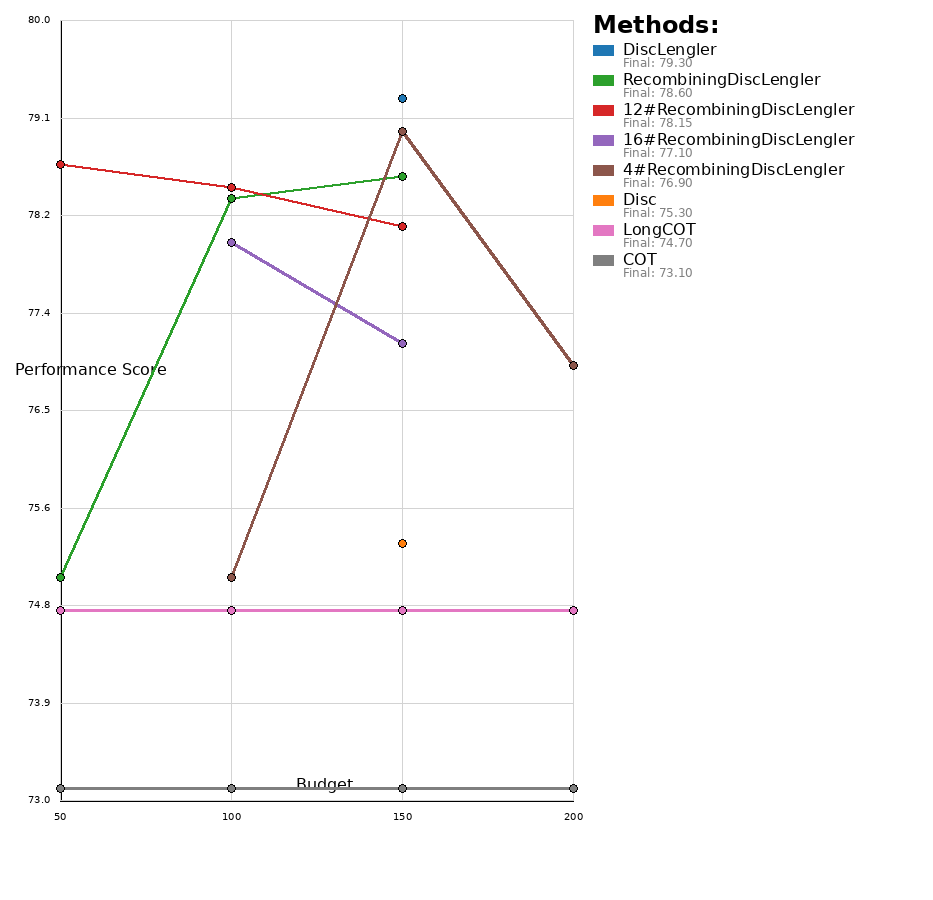}
\caption{\label{fig:svamp}Transfer of pre-prompts optimized on full GSM8k (see caption of \cref{fig:telonew}-Right) to SVAMP. We observe a good transfer in this context. }
\Description{Transfer of pre-prompts optimized on full GSM8k (see caption of \cref{fig:telonew}-Right) to SVAMP. We observe a good transfer in this context.}
\end{figure}

\noindent
{\em Generalization error: Random search vs other methods:}\label{sec:validrs}
Whereas \cref{tab:optimiser} shows poor results for random search (in terms of training error), we observe (\cref{rs100}, left) that in generalization, when we have downsampled the GSM8k training set, overfitting matters more, and random search becomes competitive. This is consistent with the limited overfitting predicted by Equation~(\ref{eq:devrs}).

\subsection{Combining with Self-Consistency}

In the context of problem-solving with LLMs, multiple pathways often lead to a solution. Recognizing this, \citet{wang2022self} introduced the concept of self-consistency~(SC) for LLMs. SC operates by generating a range of potential solutions with the same LLM. Diverse prompts for the problem at hand are generated by adding random parts and controlling the level of randomness via a specific parameter, $\tau$, called the {\em temperature}. Each of these generated pathways leads to diverse answers that are then aggregated by a majority voting~\cite{wang2022self}, effectively using the collective output of the model to determine the most likely correct answer.

We now examine the interaction between self-consistency, known to improve LLM performance, and pre-prompt optimization. The aim is to determine whether the enhancements observed in single-path decoding also hold in a self-consistency framework. Indeed, results presented in~\cref{tab:SC} show a clear benefit transferred from greedy decoding to self-consistency, at least with a temperature of 0.6 (a robust, reasonable value, though sensitivity analysis remains to be done).   

\section{Insight Analyses}\label{sec:analysis}
\begin{figure*}[t]
    \centering
    \begin{subfigure}{0.32\textwidth}
        \includegraphics[width=\textwidth]{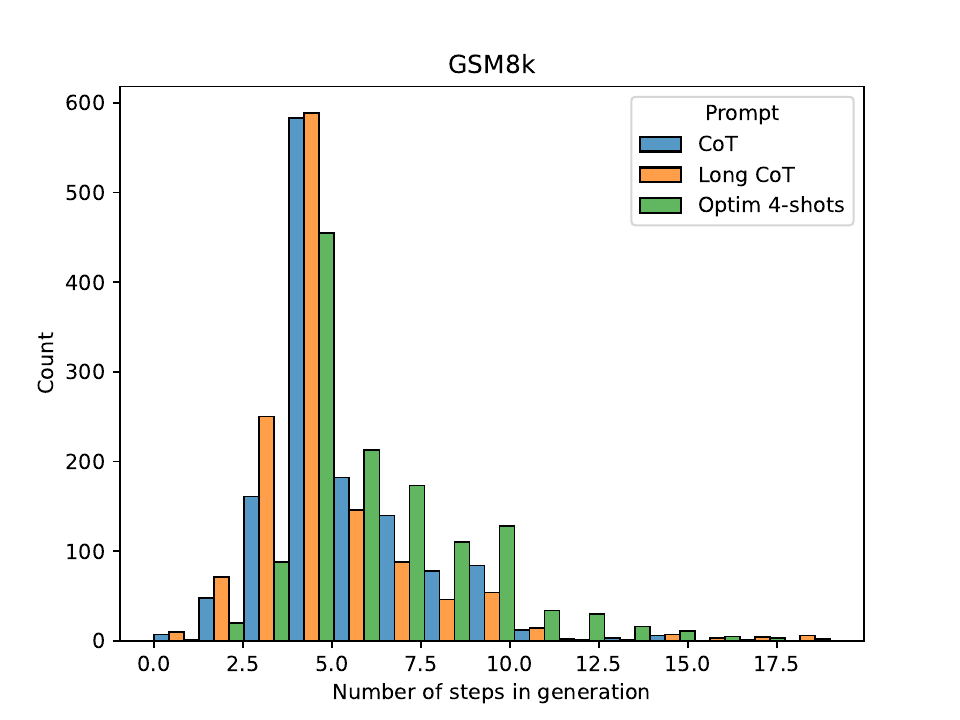}
    \end{subfigure}
    \begin{subfigure}{0.32\textwidth}
        \includegraphics[width=\textwidth]{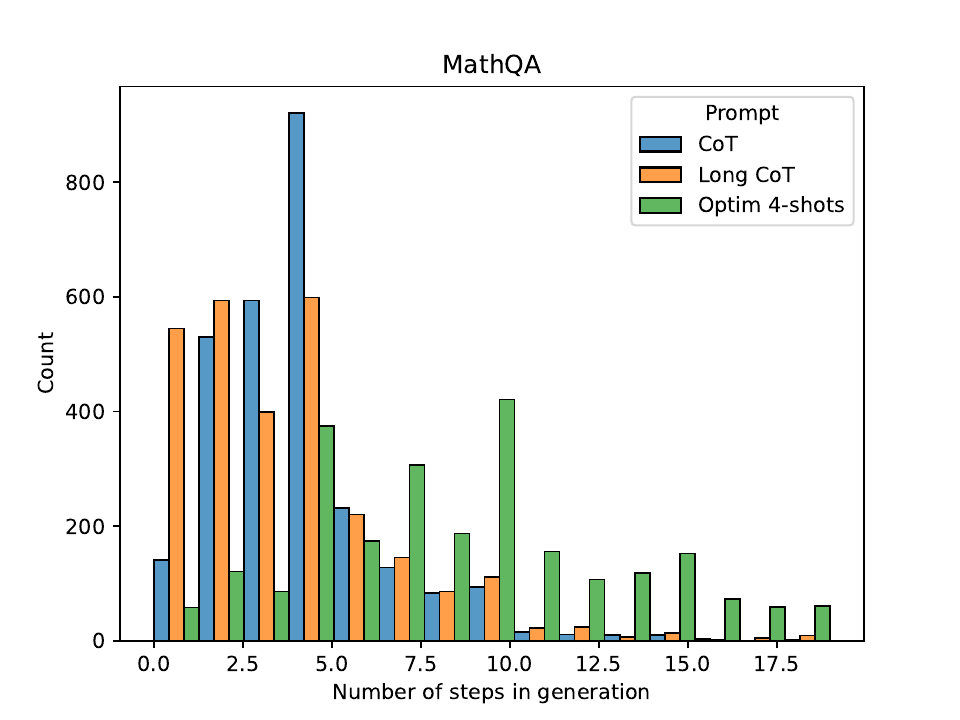}
    \end{subfigure}
    \begin{subfigure}{0.32\textwidth}
        \includegraphics[width=\textwidth]{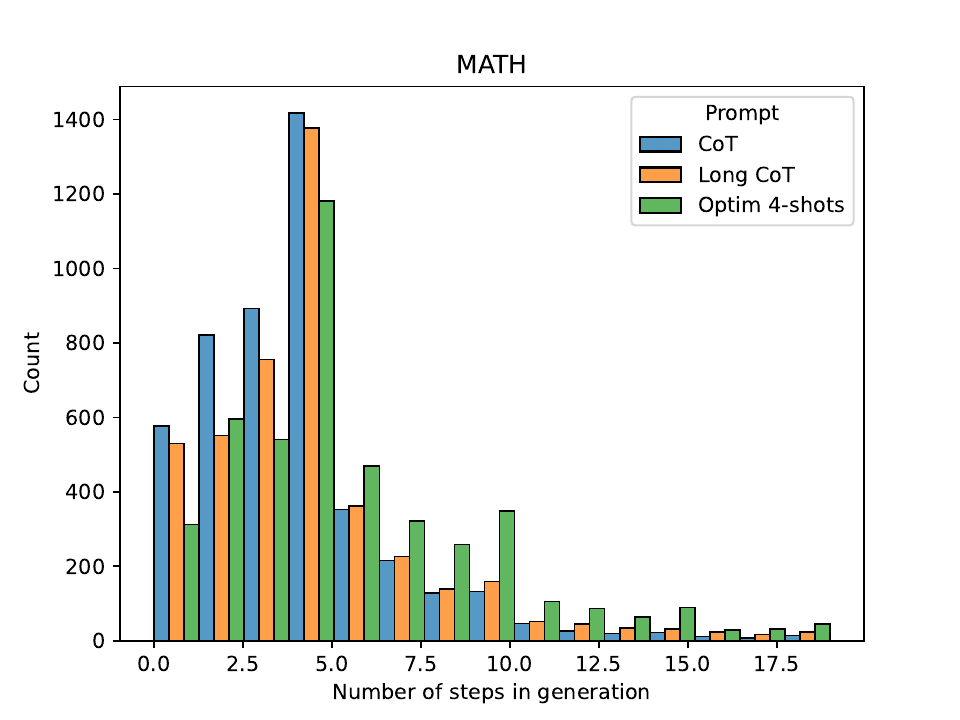}
    \end{subfigure}
    \caption{Comparison between the number of steps in the 
    CoT, Long Cot, and \EPPO\
    for GSM8k (4-shot) and MathQA (8-shot) for \llama{70B}. The variance is higher for \EPPO, with few-shot resulting in more steps in the output.}
    \Description{Comparison between the number of steps in the 
    CoT, Long Cot, and \EPPO\
    for GSM8k (4-shot) and MathQA (8-shot) for \llama{70B}. The variance is higher for \EPPO, with few-shot resulting in more steps in the output.}
    \label{fig:steps}
\end{figure*}
\begin{figure*}[t]
    \centering
    \begin{subfigure}[c]{0.3\textwidth}
        \includegraphics[width=\columnwidth]{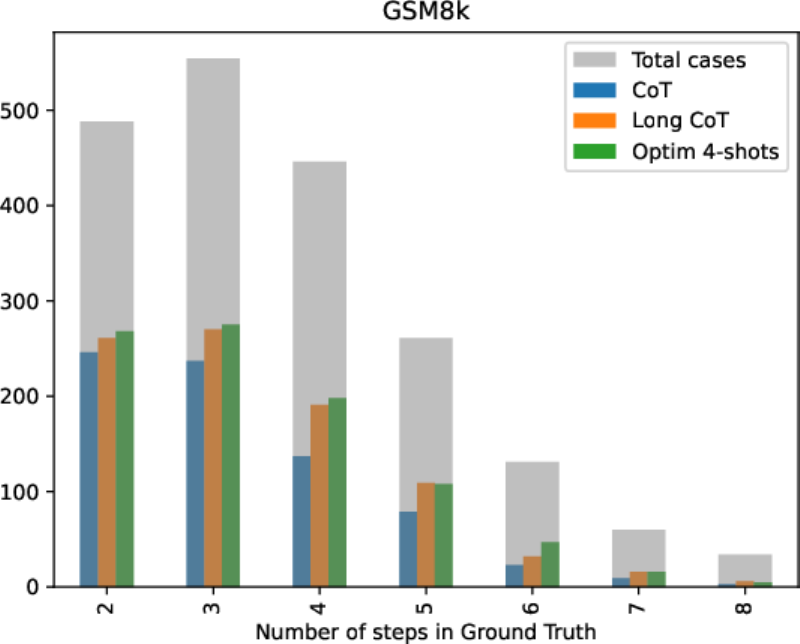}
    \end{subfigure}%
    \begin{subfigure}[c]{0.3\textwidth}
        \includegraphics[width=\columnwidth]{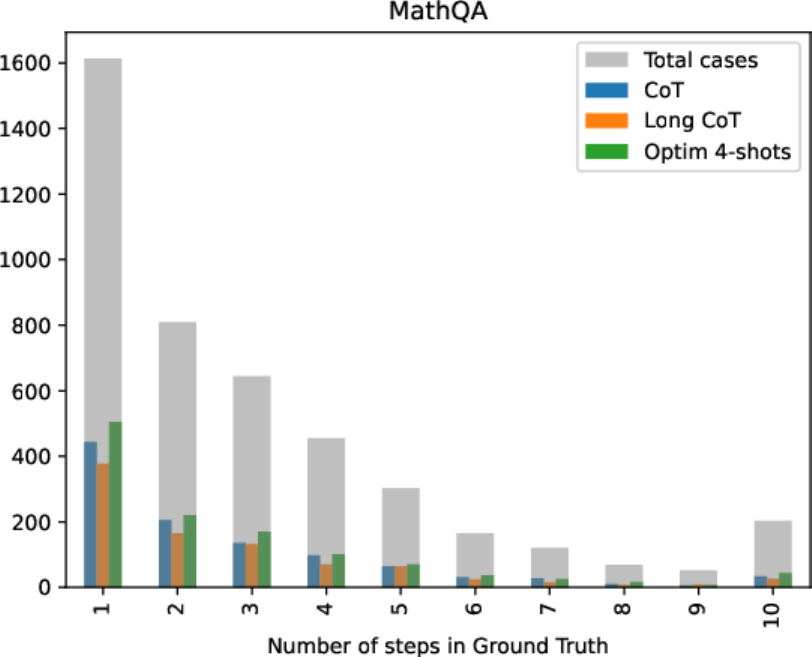}
    \end{subfigure}%
    \begin{subfigure}[c]{0.3\textwidth}
        \includegraphics[width=\columnwidth]{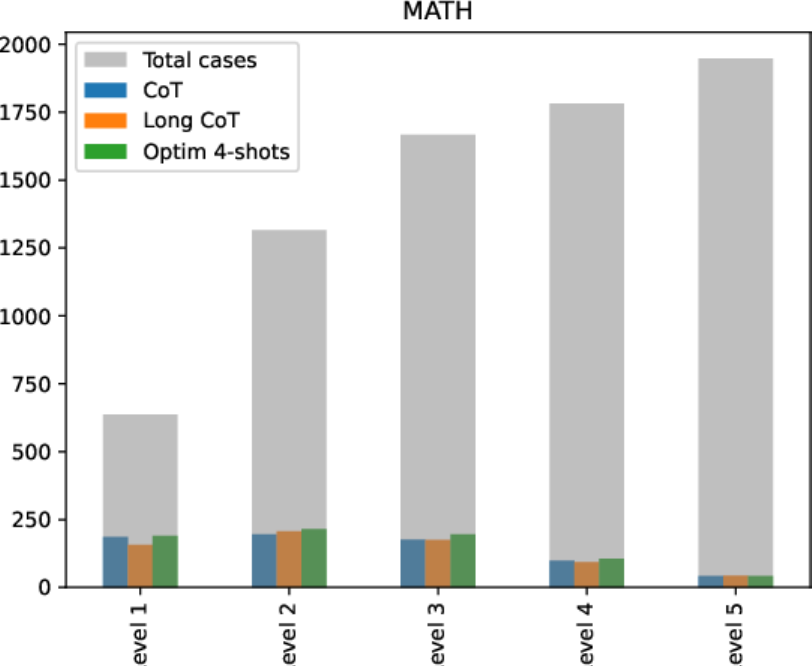}
    \end{subfigure}%
    \caption{Exact Matches on \llama{70B} for
    CoT, Long Cot, and \EPPO.
    X-axis: gold number of steps, i.e., number of steps in the ground truth or Level of the example for MATH. The x-axis thus measures the complexity of problems. \EPPO\ performs best in most cases. }
    \label{fig:complexity}
    \Description{Exact Matches on \llama{70B} 
    CoT, Long Cot, and \EPPO. 
    X-axis: gold number of steps, i.e., number of steps in the ground truth or Level of the example for MATH. The x-axis thus measures the complexity of problems. \EPPO\ performs best in most cases. }
\end{figure*}

%
\textbf{Effect of few-shot on LLMs generation:} \Cref{fig:steps} illustrates the number of steps generated by \llama{70B} across different tasks. Because the number of steps to derive a solution is hard to define, we use a simple proxy: the number of sentences generated by the LLM to derive the answers.  Contrary to baselines CoT and Long CoT, few-shot learning by \EPPO\ produces more diverse outputs in terms of the number of steps. In particular, optimized few-shot pre-prompts tend to provide answers containing more steps. This increased number of steps might help provide better answers to multi-step examples (\cref{fig:complexity}).

\noindent\textbf{Operations on optimized pre-prompts:} Here, we evaluate the robustness of the optimized few-shot prompts under various transformations: example permutations, removals, and combinations. Through these experiments, we seek to gain deeper insights into how a good few-shot pre-prompt is constructed. We perform this evaluation on GSM8k and MathQA. In this case, MathQA has been downsampled to 1000 samples to accelerate the evaluation.

\begin{figure}[h]
    \centering
    \begin{subfigure}[b]{0.4\textwidth}
    \centering
    \includegraphics[width=\columnwidth]{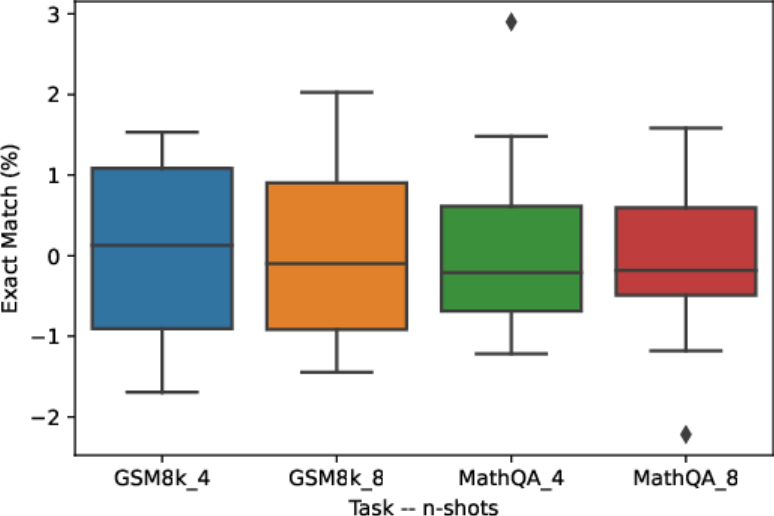}
    \caption{\llama{70B}}\label{fig:permut70}
    \end{subfigure}%
    \begin{subfigure}[b]{0.4\linewidth}
        \centering
        \includegraphics[width=\columnwidth]{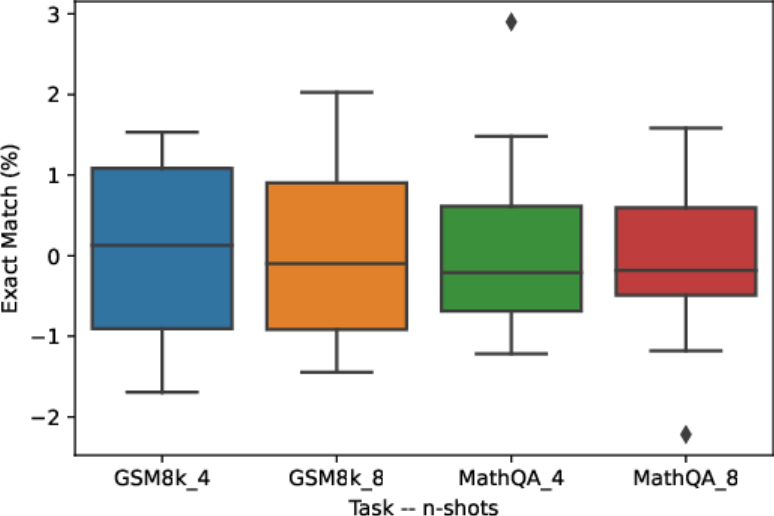}
        \caption{\llama{7B}}
    \label{fig:permut7}
    \end{subfigure}
    \caption{Differences of Exact Matches on \llama{7B} and \llama{70B} between the original pre-prompt returned by \EPPO\ and 10 randomly permuted variants: The order is not very important.
    \label{fig:permut}}
    \Description{Differences of Exact Matches on \llama{7B} and \llama{70B} between the original pre-prompt returned by \EPPO\ and 10 randomly permuted variants: The order is not very important.}
 \end{figure}

\noindent\textbf{Robustness against permutation:} \Cref{fig:permut} presents an analysis of how the performance fluctuates with changes in the order of few-shot prompts. To conduct this analysis, after running \EPPO, we randomly modified the order of the prompts in the optimized pre-prompts ten times and observed the impact on the model performance in terms of Exact Matches. 

We found that random permutations of the prompts can lead to minor variations in the results, typically around 1.

\begin{figure*}[t]
    \centering
    \begin{subfigure}{0.24\textwidth}
        \includegraphics[width=\textwidth]{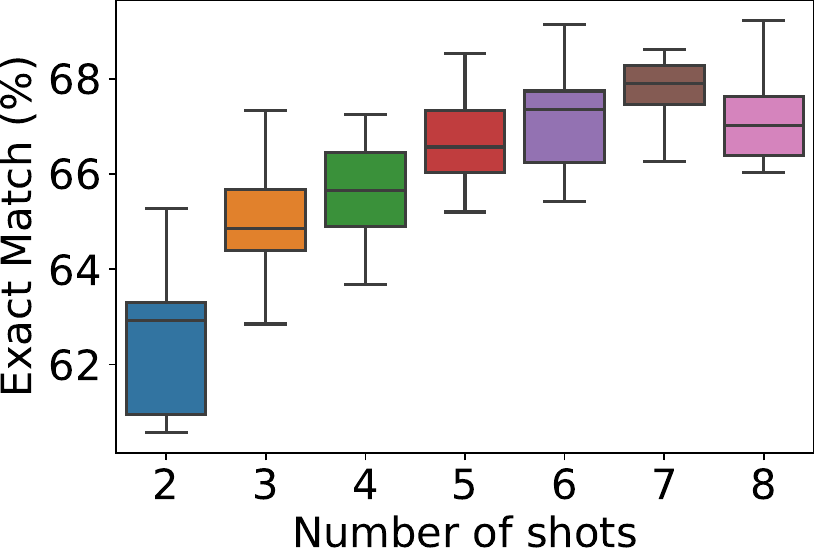}
        \subcaption{GSM8k 8-shot.}
    \end{subfigure}
    \begin{subfigure}{0.24\textwidth}
        \includegraphics[width=\textwidth]{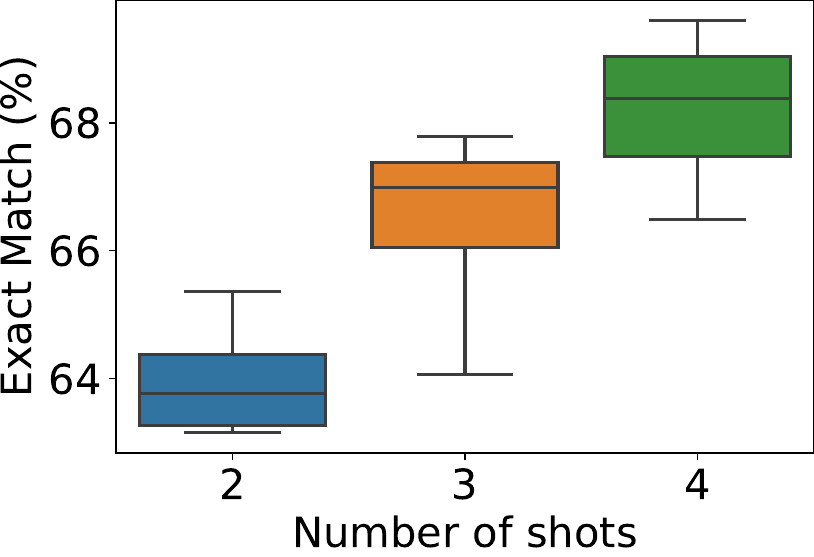}
        \subcaption{GSM8k 4-shot.}
    \end{subfigure}
    \begin{subfigure}{0.24\textwidth}
        \includegraphics[width=\textwidth]{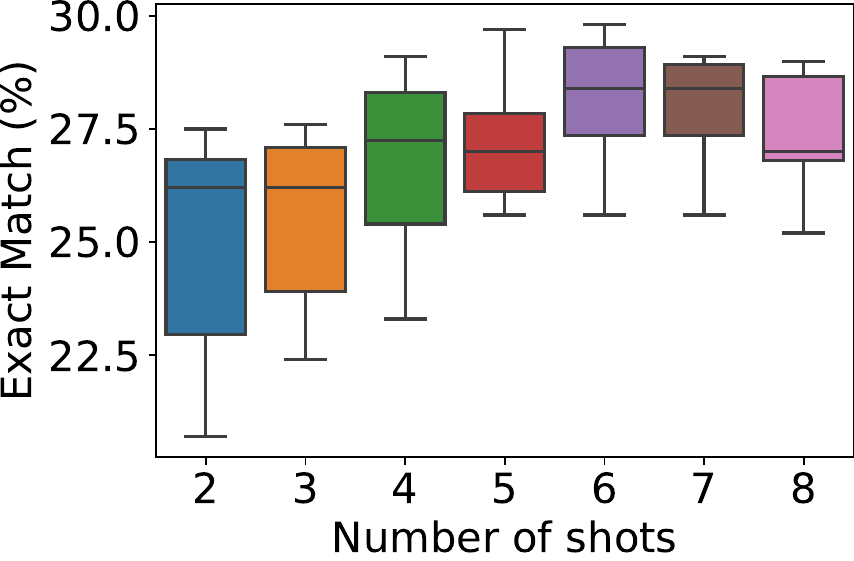}
        \subcaption{Math-QA 8-shot.}
    \end{subfigure}
    \begin{subfigure}{0.24\textwidth}
        \includegraphics[width=\textwidth]{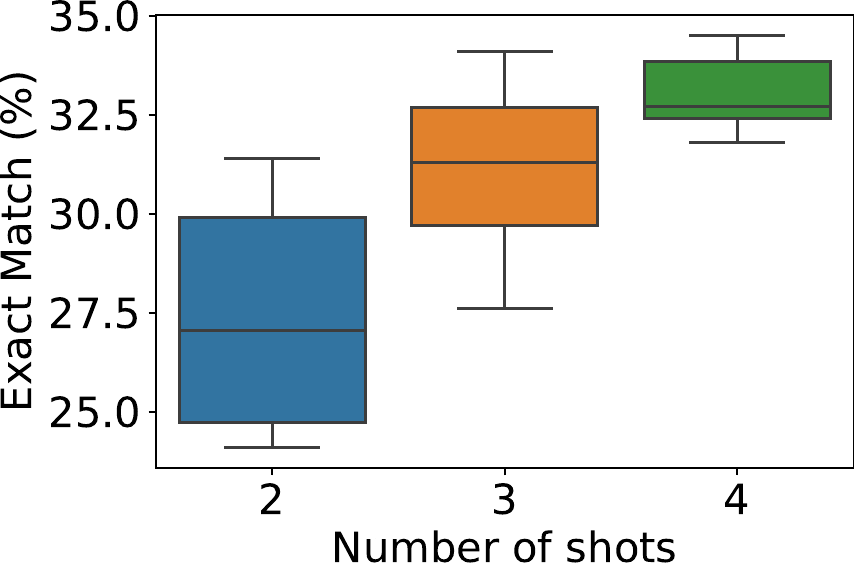}
        \subcaption{Math-QA 4-shot.}
    \end{subfigure}
    \caption{Exact Matches for \EPPO\ on {\bf \llama{70B}} after randomly reducing the optimized few-shot to $k$ shots (x-axis). Box-plots represent the distribution under 10 random permutations (as in \cref{fig:permut}).
    Reducing 8-shot to 6 or 7 does not decrease the performance, whereas any reduction does for 4-shot. }
    \label{fig:Supp}
    \Description{Exact Matches for \EPPO\ on {\bf \llama{70B}} after randomly reducing the optimized few-shot to $k$ shots (x-axis). Box-plots represent the distribution under 10 random permutations (as in \cref{fig:permut}).
    Reducing 8-shot to 6 or 7 does not decrease the performance, whereas any reduction does for 4-shot. }
\end{figure*}

\begin{figure}[!htpb]
    \centering
    \begin{subfigure}{0.23\textwidth}
        \includegraphics[width=\textwidth]{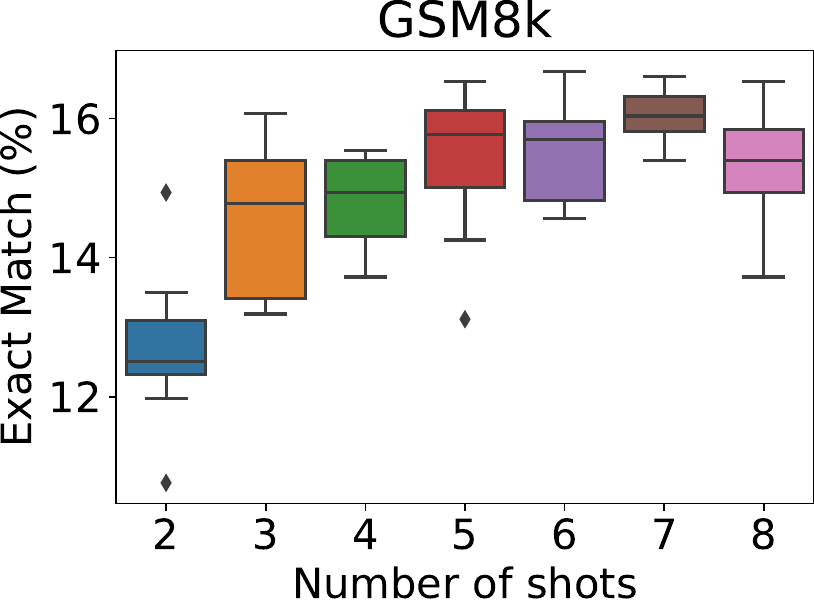}
        \subcaption{GSM8k 8-shot.}
    \end{subfigure}
    \begin{subfigure}{0.23\textwidth}
        \includegraphics[width=\textwidth]{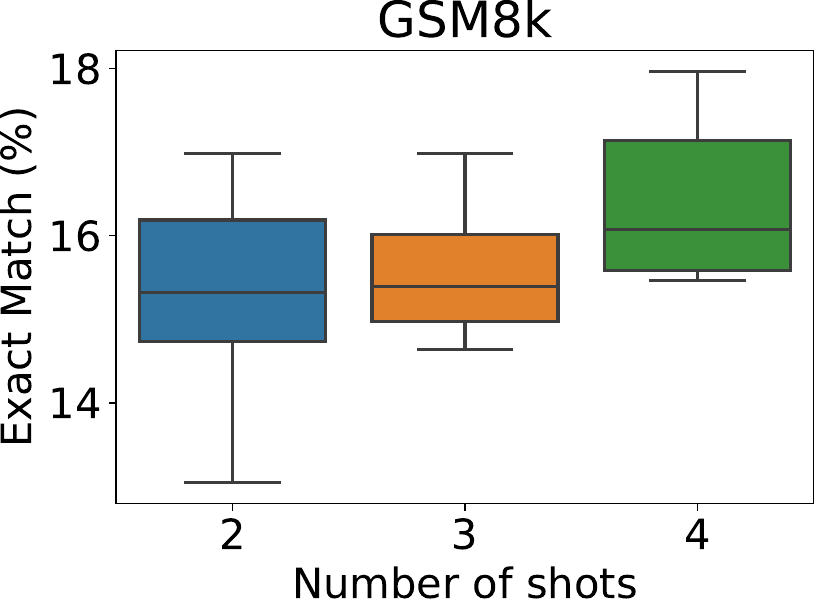}
        \subcaption{GSM8k 4-shot.}
    \end{subfigure}
    \begin{subfigure}{0.24\textwidth}
        \includegraphics[width=\textwidth]{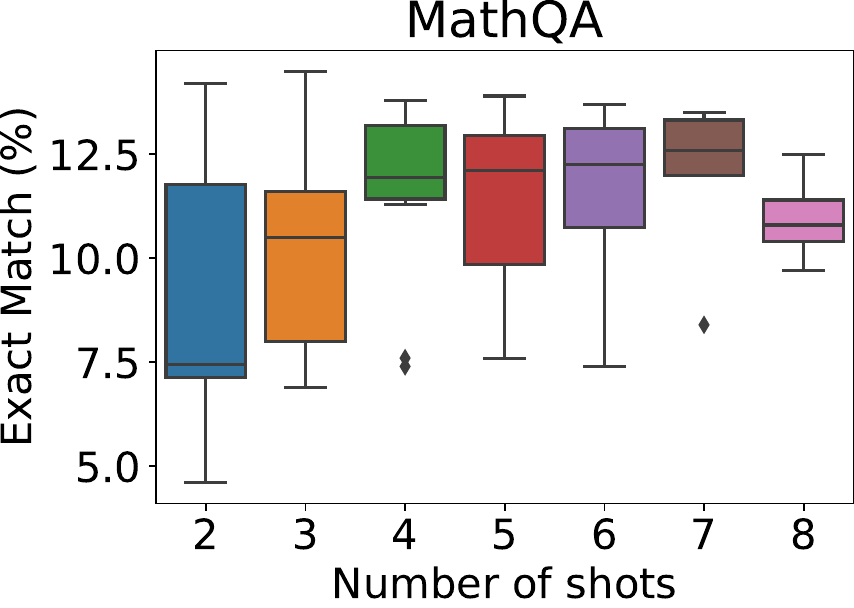}
        \subcaption{Math-QA 8-shot.}
    \end{subfigure}
    \begin{subfigure}[b]{0.23\textwidth}
        \includegraphics[width=\textwidth]{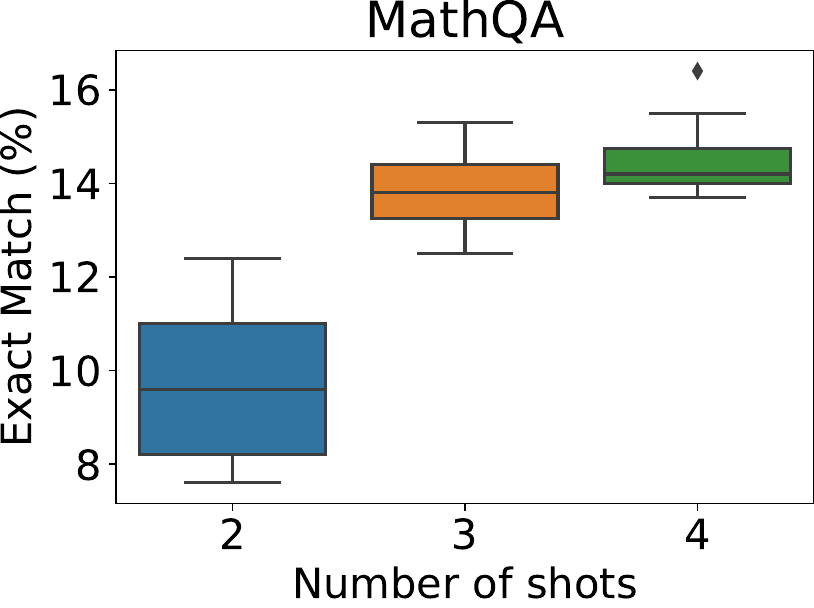}
        \subcaption{Math-QA 4-shot.}
    \end{subfigure}
    \caption{Exact Matches for \EPPO\ on {\bf \llama{7B}} after randomly reducing the optimized few-shot to $k$ shots (x-axis). Box-plots represent the distribution under 10 random permutations (as in \cref{fig:permut}).
    Reducing 8-shot to 6 or 7 does not decrease the performance, whereas any reduction does for 4-shot. }    
   \label{redper7b}
   \Description{Exact Matches for \EPPO\ on {\bf \llama{7B}} after randomly reducing the optimized few-shot to $k$ shots (x-axis). Box-plots represent the distribution under 10 random permutations (as in \cref{fig:permut}).
    Reducing 8-shot to 6 or 7 does not decrease the performance, whereas any reduction does for 4-shot. }
\end{figure}

\noindent\textbf{8-shot are more resilient to prompt removal thanks to redundancy.} To assess the sensitivity of optimized pre-prompts to removal, we randomly remove prompt examples from the few-shot pre-prompts~(\Cref{fig:Supp} for \llama{70B} and \Cref{redper7b} for \llama{7B}). The following pattern is observed in both cases: High-performing 4-shot pre-prompts lead to a significant drop in performance when some prompts are removed, as shown in~\cref{fig:Supp}-(b) and -(d). In contrast, 8-shot pre-prompts, which initially perform almost on par with 4-shot, demonstrate a higher tolerance to removal. These pre-prompts can be condensed in 6 or 7 shots without a notable decrease in performance, as indicated in~\cref{fig:Supp}-(a) and -(c). This suggests a redundancy within the elements of the 8-shot pre-prompts. This phenomenon is also visible in the MathQA dataset, despite the multiplicity of categories of questions. 

\begin{figure}[t]
    \centering
    \includegraphics[width=0.5\columnwidth]{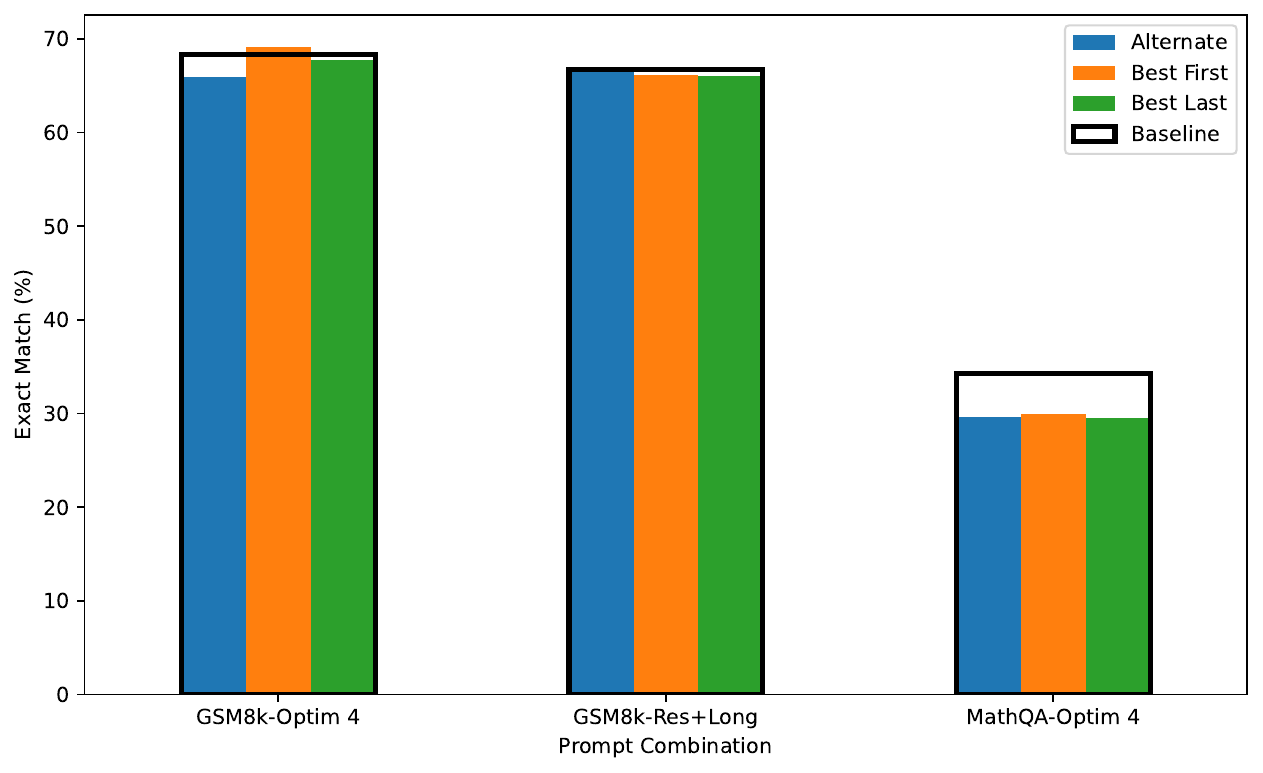}
    \caption{On \llama{70B}, test of fusion of two non-intersecting 4-shot pre-prompts obtained by \EPPO\ (see text for description of legends).
    Overall,  fusion is ineffective, and adding more shots is risky.}
    \label{fig:add_combination}
    \Description{On \llama{70B}, test of fusion of two non-intersecting 4-shot pre-prompts obtained by \EPPO\ (see text for description of legends)}
\end{figure}

\noindent\textbf{Composition of prompt:} Adding examples from an already good few-shot prompt seems like a straightforward path toward building an effective few-shot prompt. As depicted in \cref{fig:add_combination}, two strong 4-shot pre-prompts are added together to produce an 8-shot pre-prompt. Three combination strategies have been tested, and compared to the {\bf baseline} (performance of the best): 
\begin{inparaenum}[(a)]
    \item \textbf{Best first}: the most effective of the two 4-shot pre-prompts is prepended first; 
    \item \textbf{Best last}: the worst of the two 4-shot pre-prompts is prepended first; 
    \item \textbf{Alternate}: examples of both pre-prompts are interleaved.
\end{inparaenum}
Our findings reveal a slight preference of the LLM for the ``best first'' strategy, aligning with previous results~\cite{zhao2021calibrate}. However, we observe that naïvely combining pre-prompts usually degrades performance. 

\noindent\textbf{What constitutes a good pre-prompt?} We observe that chosen examples (see ``Best Few-Shot Prompt'' in the appendix
) frequently include: 
\begin{inparaenum}[(a)]
    \item Long, multi-step examples with long and intricate questions. This is consistent with some classical heuristics, such as Long CoT~\cite{fu2022complexity}.
    \item Examples which link previous steps, especially the quantities, and start by stating each element of the question very literally.  
    \item Examples which are effective while being buggy in terms of justification. We observe that the style of answer matters more than its correctness. Examples of effective incorrect rationales can be found in the MATH task few-shot.
\end{inparaenum}
However, these heuristics appear to be less effective as the difficulty of the task increases, as for MATH.

\section{Discussion}\label{sec:discussion}
{\bf{Computational cost.}} To show the effectiveness of \EPPO\ and explore further the space of examples selection in few-shot pre-prompting, we use a relatively large Demonstration dataset of around 1000 samples. Still, the computational cost of \EPPO\ is reasonable (on the downsampled GSM8k, $\approx$3h of 64 V100 GPU for \llama{70B}, less than one hour for \llama{7B} with float16 precision and a budget of 50) compared to that of training LLaMA2. Our method is tractable, in particular thanks to a well-chosen optimization method. Moreover, it has a much smaller memory footprint than approaches based on backprop (i.e., fine-tuning), as there is no backward pass involved. Moreover, a few-shot pre-prompt provided by humans can be used as an initial point for the optimization algorithm. 

\noindent
{\bf{Required data.}}
\EPPO\ also requires less data than fine-tuning: For instance, GSM8k lacks sufficient data to do a classical fine-tuning \cite{ni2022learning,liu2023tinygsm} even for moderate sized LLMs ($\approx$1B). Also, \EPPO\ is less likely to overfit on a specific task, as indicated by the information-theoretic generalization bounds (\cref{sec:maths}), independent of model size: we get positive results even with a 70B model. 
Fine-grained token-level outputs are unnecessary here and, indeed, increase the risk of overfitting. For instance, with GSM8K, only the final result is used, unlike in fine-tuning methods, where the detailed reasoning process is needed.

\noindent
{\bf{Versatility.}} We have demonstrated the efficiency of \EPPO\ on multiple datasets with various levels of difficulty and Demonstration datasets setup. In particular, for MathQA, the demonstration dataset is noisy, often containing false, poorly explained, or incomplete rationales. This particular case demonstrates the robustness of \EPPO, which nevertheless found effective few-shot preprompts. 

\noindent
{\bf{Finding the data for our approach}} is feasible: When using a dataset as ground truth, our sources are the training data and the answers by LLaMA2 itself, so human intervention is limited. Also, our approach could be entirely integrated into an automatic A/B testing scenario. Last, \EPPO\ works when heuristic methods used by humans fail. 

\noindent
{\bf{Combination with other methods.}} Finally, our improvement can be combined with Self Consistency~\cite{wang2022self}, and the improvements are additive: The simple and cheap \EPPO\ approach could also be added on top of other methods such as fine-tuning using augmented data by the LLM~\cite{yu2023metamath,xu2023wizardlm,pang2024iterative}, bootstrapping from output of the LLM~\cite{bai2022constitutional,huang2023boosting,pang2024iterative}. 

\section{Related Works}\label{sec:rw}

In-context learning has emerged as a way to adapt LLMs to a task just from a few-shot examples. This method is particularly effective as the scale of the models increases. Essentially, it allows LLMs to adjust to new tasks without modifying their underlying parameters~\cite{brown2020language,rae2021scaling,hoffmann2022training}). As LLMs continue to grow, they show improved performance across a range of tasks and exhibit the development of new skills, a phenomenon referred to as ``emergent ability''~\citep{wei2022emergent}. At the forefront of this area is the chain-of-thoughts (CoT) reasoning approach~\citep{wei2022chain}. 
As opposed to some forms of in-context learning, CoT modifies the prompt statically for a specific downstream task (e.g., mathematical reasoning) instead of on a per-example basis: Our work fits this ``per task'' framework and differs from ``per-example'' works such as \citep{r3p1,r3p2,r3p3,r1p1,r1p2,r1p3,fernando24promptbreeder}. 
This method, which prompts LLMs to articulate intermediate reasoning steps, has been refined through different techniques~\cite{wang2022self,fu2022complexity,2022aLewkowyczMinerva,jiang2023resprompt}. Despite its effectiveness, CoT prompting often faces challenges with complex multi-step tasks~\cite{fu2022complexity,zhou2022least,jiang2023resprompt}.  Various approaches have been explored to handle this challenge. For instance, \citet{wang2022self} introduced self-consistency, which involves aggregating LLMs' reasoning across multiple reasoning paths. \citet{zhou2022least} proposed a ``least to most'' prompting strategy, which breaks down problems into sub-problems and solves them sequentially. \citet{jiang2023resprompt} developed Residual prompting, which follows a reasoning graph and incorporates residual connections into the prompt design. Finally, \cite{gao2023pal} introduced PAL, a method that prompts LLMs to generate programs that solve the problem, with the solution obtained by executing these programs.\\
There are various methods for selecting examples for the prompts. This task can be very challenging due to the instability. Multiple works have shown that the performance is sensitive to prompt, task, dataset, and model changes~\cite{lu2021fantastically,zhao2021calibrate}. This sensitivity contributes to the complexity of prompt engineering, making it a somewhat arduous task. In particular, prompt engineering for LLMs often relies on community-wide collective trial-and-error efforts (there is even a prompt marketplace named PromptBase). Despite these challenges, a line of work addresses this issue by introducing retrieval into few-shot learning~\cite{zhang2022automatic,wu2023openicl,liu2021makes}. In this context, each few-shot examples are retrieved dynamically at test time for each presented example, meaning that a particular CoT is retrieved following some similarity measure on the test example and the retrieved few-shot CoT. Our work, however, shows that a single retrieval for the task is effective, robust, and transferable.%

\EPPO\ focuses on pre-prompt optimization by example prompt selection for a given task and on multi-step reasoning, searching in large datasets to demonstrate diverse ways of answering the questions.
Orthogonal to this approach, several methods~\cite{yu2023metamath,xu2023wizardlm,pang2024iterative,sessa2024bond} involve sampling multiple generations from the LLM and using bootstrapping techniques to iteratively refine and improve the model by updating its weights. These approaches are complementary to \EPPO, and they could enhance one another. Specifically, \EPPO\ could support the sampling phase by incorporating the few-shot examples it identifies, creating a more robust population of generations to bootstrap and potentially guiding the model toward even higher-quality outputs.
\change{Complementary to \EPPO\, Promptbreeder~\cite{fernando24promptbreeder} leverages LLMs to evolve the structure and the content of the prompts, whereas \EPPO\ optimizes the selection of static few-shot examples.
}

\section{Conclusion}\label{sec:conclu}
This work focused on a key component of LLMs, namely, few-shot optimization.
We showed that \change{we can outperform brute-force optimization of examples in spite of not working at a per-example level}, thanks to using relevant combinatorial optimization methods. Limiting the generalization error is crucial for obtaining such results.
Our mathematical results based on information theory predict a generalization error that \begin{inparaenum}[(i)] \item decreases as the volume of data increases  (unsurprisingly, but this is not so clear in the case in case of training on fine \change{grained} data/with next-token prediction); \item decreases when using more parallel evolutionary methods (consistently with~\cite{arxivtelo}) or random search, and \item increases with the budget (consistently with early stopping methods). 
For typical orders of magnitude, consistent with \cite{arxivtelo,liu2023tinygsm,codegenreasoning}, we obtain better generalization bounds, \ie\ reduced exploitative behavior and reduced overfitting, when using our method compared to classical fine-tuning.\end{inparaenum}

We experimentally verify each of these predictions.
Additionally, our results are robust across different contexts, such as greedy decoding LLMs versus voting mechanisms: the benefits of voting and those of few-shot optimization effectively add up. We observe a positive transfer to various contexts.

This cannot be reproduced by simply adding many random shots. In the optimized case, we observe a decrease in performance as we add shots, and the best prompts are not necessarily the longest. Also, a detailed analysis shows that few-shot optimization \change{few-shot optimization leads to a more consistent generation of reasoning steps across examples.} %

\section*{Acknowledgments}
The authors thank the reviewers for their insightful and constructive feedback, which has substantially improved the quality of this work. AL acknowledges support from the French National Research Agency (ANR) under Grant No. ANR-23-CPJ1-0099-01.%

\bibliographystyle{ACM-Reference-Format}

\bibliography{refs}

\def\checklist{
\section{Reproducibility Checklist}
Includes a conceptual outline and/or pseudocode description of AI methods introduced: \textbf{yes}
\par Clearly delineates statements that are opinions, hypotheses, and speculation from objective facts and results: \textbf{yes}
\par Provides well-marked pedagogical references for less-familiar readers to gain the background necessary to replicate the paper: \textbf{yes}\\ 
\par Does this paper make theoretical contributions? \textbf{no}\\
\par Does this paper rely on one or more datasets? \textbf{yes}\\
\par A motivation is given for why the experiments are conducted on the selected datasets: \textbf{yes}
\par All novel datasets introduced in this paper are included in a data appendix. \textbf{NA}
\par All datasets drawn from the existing literature (potentially including authors’ own previously published work) are accompanied by appropriate citations: \textbf{yes}
\par All datasets that are not publicly available are described in detail, with an explanation of why publicly available alternatives are not scientifically satisfying: \textbf{NA}\\
\par Does this paper include computational experiments? \textbf{yes}\\
\par Any code required for pre-processing data is included in the appendix: \textbf{no}
\par All source code required to conduct and analyze the experiments is included in the code appendix. \textbf{no}
\par All source code required for conducting and analyzing the experiments will be made publicly available upon publication of the paper with a license that allows free usage for research purposes. \textbf{no}
\par All source code implementing new methods has comments detailing the implementation, with references to the paper where each step comes from \textbf{no}
\par If an algorithm depends on randomness, then the method used for setting seeds is described in a way sufficient to allow replication of results: \textbf{no}
\par This paper specifies the computing infrastructure used for running experiments (hardware and software), including GPU/CPU models; amount of memory; operating system; names and versions of relevant software libraries and frameworks: \textbf{no}
\par This paper formally describes the evaluation metrics used and explains the motivation for choosing these metrics. \textbf{yes}
\par This paper states the number of algorithm runs used to compute each reported result. \textbf{no}
\par Analysis of experiments goes beyond single-dimensional summaries of performance (e.g., average; median) to include measures of variation, confidence, or other distributional information. \textbf{no}
\par The significance of any improvement or decrease in performance is judged using appropriate statistical tests (e.g., Wilcoxon signed-rank). \textbf{no}
\par This paper lists all final (hyper-)parameters used for each model/algorithm in the paper’s experiments. \textbf{yes}
\par This paper states the number and range of values tried per (hyper-) parameter during the development of the paper, along with the criterion used for selecting the final parameter setting. \textbf{no}

\FloatBarrier
}
%
\appendix
\onecolumn

\appendix
\section{Generalization risk bounds under limited feedback}
\label{sec:theory}

%
This appendix provides a simple, \emph{model-agnostic} bound on the probability that the empirical evaluation of the pre-prompt returned by \EPPO\ deviates from its expected performance. Our analysis does \emph{not} rely on gradient-based learning theory, nor on capacity measures of the underlying LLM; instead, it isolates the \emph{selection effect} induced by comparing multiple candidates. In particular, the bound depends only on \begin{inparaenum}[(i)] \item the optimizer feedback arity~$\kappa$ and \item the number of optimization iterations~$b$, through the number of candidates that can be selected.\end{inparaenum}

\subsection{Deviation event and base concentration assumption}
\label{app:base}

\noindent{\textbf{Setup.}}
Let $\mathcal{D}$ denote the demonstration pool and let a \emph{pre-prompt} $p$ be an element of $\mathcal{D}^s$ (typically, an array of $s$ indices pointing to demonstrations). Let $L(p)$ denote the (unknown) expected loss of the LLM equipped with pre-prompt~$p$ on the population distribution, and let $\widehat{L}_{\mathcal{T}}(p)$ denote its empirical loss on the evaluation set $\mathcal{T}$ (in our implementation, the full dataset used by the {\em compare step} to score or compare candidates).

For $\varepsilon>0$, define the deviation event
\begin{equation}\label{eq:app_err}
err_\varepsilon(p)
\;=\;
\Big\{\big|L(p) - \widehat{L}_{\mathcal{T}}(p)\big| > \varepsilon\Big\}.
\end{equation}

\noindent\textbf{Hypothesis H1 (single-candidate deviation bound).}
We assume that for any fixed pre-prompt $p$ and any $\varepsilon>0$ there exists a quantity $\delta_{1,\varepsilon}\in(0,1]$ such that
\begin{equation}
\label{eq:app_h1}
\Pr\!\big(err_\varepsilon(p)\big)\;\le\;\delta_{1,\varepsilon}.
\end{equation}
This is a standard concentration assumption: it bounds the probability that the empirical loss of a \emph{fixed} candidate deviates from its expectation by more than~$\varepsilon$. \\

\noindent\textbf{Example (bounded losses, Hoeffding).}
Assume $\widehat{L}_{\mathcal{T}}(p)$ is an average of $T$ independent terms taking values in $[0,1]$. Then Hoeffding's inequality yields
\begin{equation}\label{eq:devh}
\Pr\!\big(err_\varepsilon(p)\big) \;\le\; 2\exp(-2T\varepsilon^2),
\end{equation}
so Hypothesis~H1 holds with $\delta_{1,\varepsilon}=2\exp(-2T\varepsilon^2)$.

\subsection{Selection over candidates: union-bound control}
\label{app:union}\label{sec:maths}

The difficulty is that the pre-prompt returned by \EPPO\ is \emph{data-dependent}: it is chosen after observing comparison feedback. A direct application of~\cref{eq:app_h1} is therefore not immediate. We control this selection effect using the classical union bound.

\begin{lemma}[Union bound for selection over a finite candidate set]
\label{lem:app_union}
Assume Hypothesis~H1 holds.
Let $\mathcal{R}=\{r_1,\dots,r_M\}$ be a finite set of pre-prompts and let $r$ be any (possibly randomized, data-dependent) element taking values in~$\mathcal{R}$.
Then for any $\varepsilon>0$,
\begin{equation}
\label{eq:app_union}
\Pr\!\big(err_\varepsilon(r)\big) \;\le\; M\,\delta_{1,\varepsilon}.
\end{equation}
\end{lemma}

\begin{proof}
We write
\[
\Pr\!\big(err_\varepsilon(r)\big)
\;\le\;
\Pr\!\Big(\exists\, c\in\mathcal{R}:\; err_\varepsilon(c)\Big)
\;\le\;
\sum_{c\in\mathcal{R}} \Pr\!\big(err_\varepsilon(c)\big)
\;\le\;
M\,\delta_{1,\varepsilon},
\]
where the second inequality is the union bound, and the last inequality uses Hypothesis~H1.
\end{proof}

\subsection{Application to \EPPO: bounds for the returned pre-prompt and the archive}
\label{app:eppo}

We now instantiate~\cref{lem:app_union} for \EPPO. Recall that \EPPO\ runs for $b$ iterations. At each iteration, the routine \texttt{Compare} returns an index in $\{1,\dots,\kappa\}$, \ie\ the feedback is $\kappa$-ary.\footnote{For $\kappa=2$, the feedback is a single comparison bit.} The algorithm maintains an archive ({\em archive}) collecting the evaluated candidates and returns one pre-prompt $r$ at the end.

\begin{theorem}[Generalization risk bounds for \EPPO\ under limited feedback]
\label{thm:app_eppo}
Assume Hypothesis~H1 holds.
Consider one run of \cref{algoOverview} with feedback arity~$\kappa$ and budget~$b$, and let $r$ be the returned pre-prompt (\cref{line:return}) and $\text{archive}$ be the set of candidates stored along the run (\cref{line:add}).
Then for any $\varepsilon>0$,
\begin{align}
\Pr\!\big(err_\varepsilon(r)\big)
&\;\le\;
\delta_{\varepsilon}^{\EPPO}
\;\stackrel{\mathrm{def}}{=}\;
\kappa^{b}\,\delta_{1,\varepsilon},
\label{eq:app_out}
\\
\Pr\!\Big(\exists\, p\in \text{Archive}:\; err_\varepsilon(p)\Big)
&\;\le\;
\delta_{\varepsilon}^{\mathrm{unif}}
\;\stackrel{\mathrm{def}}{=}\;
\Big(\kappa+\kappa^2+\cdots+\kappa^b\Big)\delta_{1,\varepsilon}
\;=\;
\frac{\kappa^{b+1}-\kappa}{\kappa-1}\,\delta_{1,\varepsilon}.
\label{eq:app_arch}
\end{align}
\end{theorem}

\begin{proof}
\proofparagraph{Step 1: bounding the number of possible outputs}
Fix all deterministic inputs to \cref{algoOverview} (training/evaluation sets, the optimizer implementation, and the budget~$b$). If the algorithm is randomized, additionally fix its random seed~$\omega$. At each of the $b$ iterations, the {\em compare step} returns one outcome among $\kappa$ possibilities. Therefore, the entire run induces a sequence of $\kappa$-ary feedback of length~$b$, hence at most $\kappa^b$ distinct feedback transcripts. For each transcript, the run is deterministic given~$\omega$ and thus determines a unique returned pre-prompt. Consequently, for fixed~$\omega$, the returned pre-prompt must belong to a set $\mathcal{R}_\omega$ of size at most~$\kappa^b$.

\proofparagraph{Step 2: applying~\cref{lem:app_union} (fixed seed)}
Conditioned on~$\omega$, we have $r\in\mathcal{R}_\omega$ with $|\mathcal{R}_\omega|\le \kappa^b$.
Applying~\cref{lem:app_union} with $M=\kappa^b$ yields
\[
\Pr\!\big(err_\varepsilon(r)\mid \omega\big)\;\le\;\kappa^b\,\delta_{1,\varepsilon}.
\]
\proofparagraph{Step 3: removing the conditioning (randomized runs)}
If $\omega$ is random, taking expectations over~$\omega$ gives
\[
\Pr\!\big(err_\varepsilon(r)\big)
=
\mathbb{E}_\omega\!\left[\Pr\!\big(err_\varepsilon(r)\mid\omega\big)\right]
\;\le\;
\kappa^b\,\delta_{1,\varepsilon},
\]
which proves~\cref{eq:app_out}.

\proofparagraph{Step 4: uniform bound over the archive}
Across the run, at iteration $t\in\{1,\dots,b\}$, at most $\kappa^t$ distinct candidates can be \emph{considered} as a function of the first $t$ feedback outcomes (again fixing~$\omega$).
Summing over iterations yields at most $\kappa+\kappa^2+\cdots+\kappa^b$ candidates overall.
Applying \cref{lem:app_union} to the set of all candidates that can enter the archive gives~\eqref{eq:app_arch}.
\end{proof}

\noindent\textbf{Interpretation (selection effect).}
The bounds in~\cref{thm:app_eppo} depend on the number of candidates that the procedure can effectively compare and select, rather than on the size of the ambient pre-prompt space $|\mathcal{D}|^s$. This isolates the \emph{selection-induced} generalization risk: as the budget~$b$ increases, more candidates can be adaptively explored, thereby increasing the probability that at least one candidate appears spuriously good on the evaluation set.

\subsection{Instantiating H1 and scaling with dataset size}
\label{app:scaling}

When Hypothesis~H1 is instantiated via Hoeffding's inequality~\cref{eq:devrs}, \cref{thm:app_eppo} implies
\begin{equation}\label{eq:app_bigdev}
\delta^{\EPPO}_\varepsilon
\;\le\;
2\,\kappa^b \exp(-2T\varepsilon^2),
\end{equation}
and equivalently, for a target risk level $\delta^{\EPPO}_\varepsilon\in(0,1)$,
\begin{equation}
\label{eq:app_bigdev2}
\varepsilon
\;\le\;
\sqrt{\frac{\ln\!\big(\tfrac{2\kappa^b}{\delta^{\EPPO}_\varepsilon}\big)}{2T}}.
\end{equation}
These expressions quantify the trade-off between the evaluation set size~$T$ (which tightens $\delta_{1,\varepsilon}$) and the optimization budget~$b$ (which enlarges the effective candidate set via selection).

\subsection{Discussion: generalization bounds vs.\ optimization accuracy}
\label{app:discussion}

A small upper bound on the deviation probability does not imply that the method finds \emph{high-performing} candidates; it only controls the risk that empirical performance is overly optimistic. For instance, consider a form of random search that samples candidates independently and selects the best among $b$ evaluated pre-prompts. In that case, the selection set has size $b$, yielding
\begin{equation}
\label{eq:devrs}
\Pr\!\big(err_\varepsilon(r)\big)\;\le\; b\,\delta_{1,\varepsilon},
\end{equation}
which can be smaller than the $\kappa^b\,\delta_{1,\varepsilon}$ bound for evolutionary selection. However, random search is typically far less effective at identifying empirically strong pre-prompts. This highlights a basic tension between \begin{inparaenum}[(i)] \item exploring many candidates to improve empirical performance and \item controlling selection-induced generalization risk. \end{inparaenum} A natural direction is to study intermediate strategies that increase population-level diversity (larger~$\kappa$ with fewer iterations, early stopping, or stronger validation protocols) to balance these objectives.

\section{Datasets and hyperparameters}\label{sec:datasetshyper}
\FloatBarrier
\begin{table}[h]
    \centering
    \caption{Information on datasets.}\label{tab:data_info}
    \rowcolors{2}{gray!25}{white}
    \begin{adjustbox}{width=\columnwidth}
    \begin{tabular}{lcccc}
        \toprule
        Dataset & \#Test Samples & \#Original Train Sample & \#Subsampled Train & \#Demonstration Sample\\
        \hline
        GSM8K & 1319 & 7473 & 400 & 1000\\
        SVAMP & 1000 & 0 & 0 & 0\\
        MathQA & 2985 & 29837 & 490 & 1000\\
        MATH & 5000 & 7500 & 520 & 800\\
        \bottomrule
    \end{tabular}
    \end{adjustbox}
\end{table}
\FloatBarrier
\section{More results on model transfer}
\FloatBarrier
\cref{tab:model_transfert} presents a counterpart of \cref{tab:results:70b,tab:results:7b} in the case of LLaMA2 chat models instead of LLaMA2.
\begin{table*}[!htpb]
    \centering
    \caption{Transfer to chat counterparts: compared to CoT or Long CoT (few-shots with long examples), we get slightly better results (on average) but the improvement is moderate. Contrary to non chat models, 70B prompts can transfer to the 7B chat. 7B chat models are distilled from the 70B chat: this might explain this behavior changing}
    \label{tab:model_transfert}
    \begin{tabular}{l|c|cccc}
    \toprule
    \multicolumn{1}{c}{Model} & Prompt Type & GSM8k & SVAMP & MathQA & MATH \\
    \midrule
      \multirow{5}{*}{\llama{7B-chat}} & CoT & 20.2 & \textbf{57.4} & 6.4 (22.7) & 4.8\\
      & Long CoT & 16.7 & 48.2 & 5.8 (21.6) & 4.7\\
      & Resprompt & 17.3 & 50.7 & 6.2 (22.0) & -\\
      \cmidrule{2-6}
      & 70B $\rightarrow$ 7B-chat & 21.9 & 56.6 & {\bf{11.5 (24.0)}} & 4.8\\
      & 7B $\rightarrow$ 7B-chat & {\bf{24.6}} & 51.8 & 10.0 (20.6) & {\bf{5.0}}\\
      \cmidrule{1-6}
      \multirow{5}{*}{\llama{70B-chat}} & CoT & 48.1 & 75.4 & 20.2 (36.2) & {\bf{11.7}}\\
       & Long CoT & \textbf{56.6} & 77.2 & 21.7 (34.9) & 11.1\\
       & Resprompt & 52.3 & 75.5 & 21.7 (35.3) & - \\
       \cmidrule{2-6}
       & 70B $\rightarrow$ 70B-chat & 53.0 & {\bf{78.9}} & 21.2 (34.5) & {\bf{11.7}} \\
       & 7B $\rightarrow$ 70B-chat & 50.6 & 78.3 & {\bf{22.1 (35.6)}} &  {\bf{11.7}}\\
    \bottomrule
    \end{tabular}    
\end{table*}
\FloatBarrier
\section{More results on task transfer}
\FloatBarrier
\begin{table}[!htpb]
    \centering
    \caption{\llama{70B}. Comparison, on MathQA and MATH, between CoT and preprompts transferred from GSM8k. While the preprompt from CoT-GSM8k does not outperform CoT, Optim 4-GSM8k transfers reasonably well.}\label{tab:task_transfert}
    \begin{tabular}{l|cc}
    \toprule
    \multicolumn{1}{c}{Prompt Type} & MathQA & MATH \\
    \midrule
    CoT       & 25.4 (35.8) & 13.9 \\
    CoT GSM8k & 20.6 (30.7) & 13.5 \\
    Optim 4 GSM8k & {\bf{26.0 (27.3)}} & {\bf{14.0}} \\
    \bottomrule    
    \end{tabular}    
\end{table}    
\FloatBarrier

\section{Best Few-Shot Prompt}\label{sec:bestshotprompt}
\FloatBarrier
GSM8k best 8 shot examples for \llama{70B} (keep the last 4 to get the best 4 shots). Buggy rationales are highlighted in red.\\
\rule{\textwidth}{1pt}
Q: James spends 10 minutes downloading a game, half as long installing it, and triple that combined amount of time going through the tutorial. How long does it take before he can play the main game?

A: James spends 10 minutes downloading a game, half as long (10 / 2 = 5 minutes) installing it, and triple that combined amount of time (10 + 5 = 15) going through the tutorial. Therefore, he spends 10 + 5 + 3 * 15 = 10 + 5 + 45 = 60 minutes before he can play the main game. The answer is 60.

Q: Alicia had a wonderful birthday party where she got lots of presents. 10 of the presents were in small boxes. 12 of the presents were in medium boxes. A third of all the presents she is given are in large boxes. How many presents did Alicia get for her birthday?

A: Alicia got 10 presents in small boxes and 12 presents in medium boxes. A third of the presents are in large boxes, so the number of presents in large boxes is 1/3 of the total number of presents. The total number of presents is 10 + 12 + 1/3 * total number of presents = 10 + 12 + total number of presents/3 = 22 + total number of presents/3. Solving for total number of presents, we get 22 * 3 = total number of presents + total number of presents. Therefore, total number of presents + total number of presents = 66. Since this is a sum of the number of presents with itself, we can divide both sides by 2 to get total number of presents = 66 / 2 = 33. The answer is 33.

Q: Max works for a delivery company as a courier. During one working day, he can deliver a maximum of 35 packages. One week, Max delivered the maximum number of packages, only twice. On two other days, Max unloaded a total of 50 packages, and on one day he unloaded only one-seventh of the maximum possible daily performance. On the last two days, the sum of the packages transported was only fourth-fifth of the maximum daily performance. How many more packages could Max deliver during that week, if he would work at maximum performance?

A: Max works as a courier, and he can deliver a maximum of 35 packages per day. He delivered the maximum number of packages, twice. Therefore, he delivered 35 * 2 = 70 packages on those two days. On two other days, Max unloaded 50 packages, and on one day he unloaded only one-seventh of the maximum possible daily performance. So, Max delivered 50 + (35/7) = 50 + 5 = 55 packages on those three days. On the last two days, the sum of the packages transported was only fourth-fifth of the maximum daily performance. So, he delivered 35 * (4/5) = 28 packages on those two days. Adding these together, 70 (on the two days when he delivered the maximum) + 55 (on the three days when he delivered 50 and 5 packages) + 28 (on the two days when he delivered 4/5 of the maximum) = 153 packages in total. The maximum number of packages that he could have delivered is 35 * 7 = 245 packages. So, 245 - 153 = 92 more packages could have been delivered if he worked at maximum performance. The answer is 92.

Q: Remi wants to drink more water. He has a refillable water bottle that holds 20 ounces of water. That week Remi refills the bottle 3 times a day and drinks the whole bottle each time except for twice when he accidentally spills 5 ounces the first time and 8 ounces the second time. In 7 days how many ounces of water does Remi drink?

A: Remi has a refillable water bottle that holds 20 ounces. He refills the bottle 3 times a day for 7 days. 3 refills per day * 7 days = 21 refills. Each refill is 20 ounces, so he drinks 20 * 21 = 420 ounces of water. However, he spilled 5 ounces the first time and 8 ounces the second time, so he lost 5 + 8 = 13 ounces of water. Therefore, the total amount of water he drinks is 420 - 13 = 407 ounces. The answer is 407.

Q: Jenny wants to sell some girl scout cookies and has the choice of two neighborhoods to visit. Neighborhood A has 10 homes which each will buy 2 boxes of cookies. Neighborhood B has 5 homes, each of which will buy 5 boxes of cookies. Assuming each box of cookies costs \$2, how much will Jenny make at the better choice of the two neighborhoods?

A: Jenny wants to sell some girl scout cookies and has the choice of two neighborhoods to visit. Neighborhood A has 10 homes which each will buy 2 boxes of cookies. Neighborhood B has 5 homes, each of which will buy 5 boxes of cookies. Each box of cookies costs \$2. In Neighborhood A, there are 10 homes that each buy 2 boxes of cookies, so Jenny will make 10 * 2 = 20 boxes * \$2 per box = \$40. In Neighborhood B, there are 5 homes that each buy 5 boxes of cookies, so Jenny will make 5 * 5 = 25 boxes * \$2 per box = \$50. The better choice is Neighborhood B, so Jenny will make \$50. The answer is 50.

Q: Jaylen has 5 carrots and 2 cucumbers. Jaylen has twice as many bell peppers as Kristin. Jaylen has 3 less than half as many green beans as Kristin. If Kristin has 2 bell peppers and 20 green beans, how many vegetables does Jaylen have in total?

A: Jaylen has 5 carrots and 2 cucumbers. Jaylen has twice as many bell peppers as Kristin. Jaylen has 3 less than half as many green beans as Kristin. Kristin has 2 bell peppers and 20 green beans. Half of 20 is 10, and 3 less than half is 7. So Jaylen has 7 green beans. Jaylen has twice as many bell peppers as Kristin, so he has 2 * 2 = 4 bell peppers. In total, Jaylen has 5 carrots + 2 cucumbers + 4 bell peppers + 7 green beans = 18 vegetables. The answer is 18.

Q: Kristy, a sales representative earns a basic salary of \$7.50 per hour plus a 16\% commission on everything she sells. This month, she worked for 160 hours and sold \$25000 worth of items. Her monthly budget for food, clothing, rent, transportation, bills and savings is 95\% of her total monthly earnings and the rest will be put towards insurance. How much did she allocate to insurance?

A: Kristy earns a basic salary of \$7.50 per hour. She worked for 160 hours, so her basic salary was \$7.50 * 160 = \$1200. She also earned 16\% commission on everything she sold. She sold \$25000 worth of items, so she earned \$25000 * 0.16 = \$4000 in commission. Her total monthly earnings is her basic salary plus her commission: \$1200 + \$4000 = \$5200. She allocated 95\% of her total monthly earnings to her monthly budget (food, clothing, rent, transportation, bills and savings). Therefore, her monthly budget is 95\% of \$5200, which is \$5200 * 0.95 = \$4940. The rest will be put towards insurance, so the amount she allocated to insurance is \$5200 - \$4940 = \$260. The answer is 260.

Q: Stephanie is moving into a new apartment. She needs to figure out how many pieces of silverware she should purchase. She needs spoons, butter knives, steak knives, and forks. For herself she figures 5 of each would be sufficient. But in case she has guests, she wants to have 10 extra pieces of each type. Then she realizes that if everything is clean at the same time, that she will have too many of everything and not enough room to fit them in her kitchen silverware drawer. So she decides to purchase 4 fewer spoons, 4 fewer butter knives, 5 fewer steak knives, and 3 fewer forks. How many pieces total of all the silverware is Stephanie going to buy?

A: Stephanie wants to get 5 each of each type of silverware for herself. 5 spoons + 5 butter knives + 5 steak knives + 5 forks = 20 pieces of silverware She wants 10 extra of each type for guests, 10 x 4 = 40 She decides to purchase less of each type. She was originally purchasing 15 of each type. 15 spoons - 4 less spoons = 11 spoons. 15 butter knives - 4 less = 11. 15 steak knives - 5 = 10. 15 forks - 3 less forks = 12. Stephanie is going to purchase 11 spoons, 11 butter knives, 10 steak knives, and 12 forks. 11 + 11 + 10 + 12 = 44. Stephanie is going to purchase 44 pieces of silverware. The answer is 44.\\
\rule{\textwidth}{1pt}
\FloatBarrier
MathQA best 4 shot examples for \llama{70B}\\
\rule{\textwidth}{1pt}
Question: Bucket a is 1 / 4 full of sand, and bucket b, which has one - half the capacity of bucket a, is 3 / 8 full of sand. Jana pours the contents of bucket b into bucket a. Bucket a is now filled to what fraction of its capacity?\\
Options: (a) 3 / 8 , (b) 7 / 16 , (c) 1 / 2 , (d) 5 / 8 , (e) 7 / 8

Answer: Let capapcity of a = x sand = x / 4 capacity of b = x / 2 sand = 3 x / 16 a total = x / 4 + 3 x / 16 = 7 x / 16. Therefore, the final answer is 7 / 16. This matches option (b).

Question: The s.i. On a certain sum of money for 6 years at 14 \% per annum is half the c.i. On rs.7000 for 2 years at 7 \% per annum. The sum placed on s.i. Is?\\
Options: (a) 603.75 , (b) 555.75 , (c) 569.55 , (d) 256.25 , (e) 563.23

Answer: c.i. = [ 7000 * ( 1 + 7 / 100 ) 2 - 7000 ] = ( 7000 * 11 / 10 * 11 / 10 - 7000 ) = rs.1014.3. Sum = ( 507.15 * 100 ) / ( 6 * 14 ) = rs.603.75. Therefore, the final answer is 603.75. This matches option (a).

Question: In a throw of dice what is the probability of number greater than 1\\
Options: (a) 1 / 2 , (b) 1 / 3 , (c) 1 / 5 , (d) 5 / 6 , (e) none of these

Answer: Explanation : number greater than 1 is 2,3,4,5 \& 6, so only 5 number total cases of dice = [ 1,2,3,4,5,6 ] so probability = 5 / 6. Therefore, the final answer is none of these. This matches option (e).

Question: 12 welders work at a constant rate they complete an order in 8 days. If after the first day,9 welders start to work on the other project, how many more days the remaining welders will need to complete the rest of the order?\\
Options: (a) 5 , (b) 12 , (c) 18 , (d) 24 , (e) 28

Answer: 1. We need to find out the time taken by 3 workers after day 1.2. Total no. Of wokers * total time taken = time taken by 1 worker 3. Time taken by 1 worker = 12 * 8 = 96 days 4. But on day 1 twelve workers had already worked finishing 1 / 8 of the job. So 3 workers have to finish only 7 / 8 of the job.5. Total time taken by 3 workers can be got from formula used at ( 2 ).i.e.,3 * total time taken = 96. Total time taken by 3 workers to finish the complete job is 96 / 3 = 32 days.6. Time taken by 6 workers to finish 7 / 8 of the job is 7 / 8 * 32 = 28 days. Therefore, the final answer is 28. This matches option (e).\\
\rule{\textwidth}{1pt}
\FloatBarrier
MATH best 4 shot examples for \llama{70B}\\
\rule{\textwidth}{1pt}
Problem:\\
Find all the solutions to
\[\frac{1}{x^2 + 11x - 8} + \frac{1}{x^2 + 2x - 8} + \frac{1}{x^2 - 13x - 8} = 0.\]Enter all the solutions, separated by commas.\\\\
Solution:
Let $y = x^2 - 13x - 8.$  Then we can write the given equation as
\[\frac{1}{y + 24x} + \frac{1}{y + 15x} + \frac{1}{y} = 0.\]Multiplying everything by $(y + 24x)(y + 15x)y,$ we get
\[(y + 15x)y + y(y + 24x) + (y + 24x)(y + 15x) = 0.\]This simplifies to $360x^2 + 78xy + 3y^2 = 0,$ which factors as $3(20x + y)(6x + y) = 0.$  Hence, $20x + y = 0$ or $6x + y = 0.$

If $20x + y = 0,$ then $20x + x^2 - 13x - 8 = x^2 + 7x - 8 = (x - 1)(x + 8) = 0,$ so $x = 1$ or $x = -8.$

If $6x + y = 0,$ then $6x + x^2 - 13x - 8 = x^2 - 7x - 8 = (x - 8)(x + 1) = 0,$ so $x = 8$ or $x = -1.$  Thus, the solutions are $\boxed{8,1,-1,-8}.$
\\Final Answer: The final answer is $8,1,-1,-8$. I hope it is correct.

Problem:\\
A palindrome is an integer that reads the same forward and backward, such as 3663. What percent of the palindromes between 100 and 500 contain at least one 5?\\\\
Solution:
Since none of the palindromes between 100 and 500 begin with a 5, the only place a 5 can appear is in the tens digit.  Therefore, there are 4 palindromes between 100 and 500 with a 5: 151, 252, 353, and 454.  To count the total number of palindromes, we observe that there are 4 choices to make for the first/last digit and 10 choices to make for the middle digit.  Therefore, the percentage of palindromes that contain 5 as a digit is $\frac{4}{4\cdot 10}=\boxed{10\%}$.\\
Final Answer: The final answer is $10\%$. I hope it is correct.

Problem:\\
Let $A = (1,0)$ and $B = (5,4).$  Let $P$ be a point on the parabola $y^2 = 4x.$  Find the smallest possible value of $AP + BP.$\\\\
Solution:\\
We are given two points on the coordinate plane, $A = (1,0)$ and $B = (5,4)$, and a parabola $y^2 = 4x$. We are asked to find the smallest possible value of $AP + BP$, where $P$ is a point on the parabola.
To find the smallest value of $AP + BP$, we can use the fact that the sum of the distances from two points on a parabola to a given point is constant. This means that the value of $AP + BP$ will be the same for all points on the parabola.
To find the smallest value of $AP + BP$, we can minimize the sum of the distances by finding the point on the parabola that is closest to both $A$ and $B$. This point can be found by using the fact that the tangent line to a parabola at a given point is perpendicular to the line through the focus and the point.
In this case, the focus of the parabola is $(2,0)$, and the line through the focus and point $A$ is given by the equation $y = \frac{4}{5}(x-1)$. The tangent line to the parabola at the point on the parabola closest to $A$ is perpendicular to this line, so it has the equation $y = -\frac{5}{4}(x-1)$.
The point on the parabola closest to $A$ is the intersection of these two lines, which is given by the equation $y^2 = 4x$ and $y = -\frac{5}{4}(x-1)$. Solving this system of equations, we find that the point on the parabola closest to $A$ is $(2,0)$.
Similarly, we can find the point on the parabola closest to $B$ by using the fact that the tangent line to the parabola at this point is perpendicular to the line through the focus and point $B$. This point is also $(2,0)$.
Since the point on the parabola closest to both $A$ and $B$ is $(2,0)$, the smallest possible value of $AP + BP$ is the distance from $(2,0)$ to $(1,0)$ plus the distance from $(2,0)$ to $(5,4)$. This can be calculated using the distance formula:
$AP = \sqrt{(2-1)^2 + (0-0)^2} = \sqrt{1} = 1$ and $BP = \sqrt{(2-5)^2 + (0-4)^2} = \sqrt{25} = 5$.
Therefore, the smallest possible value of $AP + BP$ is $1 + 5 = \boxed{6}$\\
Final Answer: The final answer is $6$. I hope it is correct.

Problem:\\
The mean of $5,8$ and $17$ is equal to the mean of $12$ and $y$. What is the value of $y$?
\\
Solution:
To find the mean, we add up the terms and divide by the number of terms. The mean of $5, 8$ and $17$ is $\frac{5+8+17}{3}=\frac{30}{3}=10$. We set this equal to the mean of $12$ and $y$ and get $$10=\frac{12+y}{2}\qquad\Rightarrow 20=12+y\qquad\Rightarrow 8=y.$$ The value of $y$ is $\boxed{8}$.\\
Final Answer: The final answer is $8$. I hope it is correct.

\end{document}